\documentclass[sigconf]{acmart}

\usepackage{graphicx}
\usepackage{amsmath}
\usepackage{subfigure}
\usepackage{multirow}
\usepackage{caption}
\usepackage{balance}
\AtBeginDocument{%
  \providecommand\BibTeX{{%
    \normalfont B\kern-0.5em{\scshape i\kern-0.25em b}\kern-0.8em\TeX}}}

\copyrightyear{2020}
\acmYear{2020}
\setcopyright{acmcopyright}
\acmConference[MM '20]{Proceedings of the 28th ACM International Conference on Multimedia}{October 12--16, 2020}{Seattle, WA, USA}
\acmBooktitle{Proceedings of the 28th ACM International Conference on Multimedia (MM '20), October 12--16, 2020, Seattle, WA, USA}
\acmPrice{15.00}
\acmDOI{10.1145/3394171.3413569}
\acmISBN{978-1-4503-7988-5/20/10}

\begin{document}
\fancyhead{}
\title{ATRW: A Benchmark for Amur Tiger Re-identification in the Wild}

\author{Shuyuan Li}
\affiliation{%
  \institution{Shanghai Jiao Tong University}
}
\email{shuyuanli@sjtu.edu.cn}
\author{Jianguo Li}
\authornote{Corresponding author}
\affiliation{
  \institution{Ant Group}
}
\email{jglee@outlook.com}
\author{Hanlin Tang}
\affiliation{
  \institution{Intel Corporation}
}
\email{hanlin.tang@intel.com}
\author{Rui Qian}
\affiliation{
  \institution{Shanghai Jiao Tong University}
}
\email{qrui9911@sjtu.edu.cn}
\author{Weiyao Lin}
\affiliation{
  \institution{Shanghai Jiao Tong University}
}
\email{wylin@sjtu.edu.cn}


\renewcommand{\shortauthors}{Li and Li, et al.}

\begin{abstract}
   Monitoring the population and movements of endangered species is an important task to wildlife conversation. Traditional tagging methods do not scale to large populations, while applying computer vision methods to camera sensor data requires re-identification (re-ID) algorithms to obtain accurate counts and moving trajectory of wildlife. However, existing re-ID methods are largely targeted at persons and cars, which have limited pose variations and constrained capture environments.
   This paper tries to fill the gap by introducing a novel large-scale dataset, the Amur Tiger Re-identification in the Wild (ATRW) dataset. ATRW contains over 8,000 video clips from 92 Amur tigers, with bounding box, pose keypoint, and tiger identity annotations. In contrast to typical re-ID datasets, the tigers are captured in a diverse set of unconstrained poses and lighting conditions.
   We demonstrate with a set of baseline algorithms that ATRW is a challenging dataset for re-ID.
   Lastly, we propose a novel method for tiger re-identification,
   which introduces precise pose parts modeling in deep neural networks to handle large pose variation of tigers,
   and reaches notable performance improvement over existing re-ID methods. The ATRW dataset is public available at \url{https://cvwc2019.github.io/challenge.html}
\end{abstract}




\maketitle
\vspace{-1ex}
\section{Introduction}

Wildlife conservation is critical for maintaining species biodiversity. Failure to protect endangered species on Earth may lead to imbalance ecosystems \cite{biodiversity2} and affect environmental health \cite{biodiversity}. This mission is increasingly depended on accurate monitoring of the geospatial distribution and population health of these endangered species \cite{martin2007}, especially in the face of poaching and loss of habitats. Traditional methods of attaching transmitters to wildlife are prone to sensor failure, difficult to scale to large populations, and cannot measure how the wildlife interacts with its environment.

\begin{figure*}[t]
    \centering
    \includegraphics[width=0.8\linewidth]{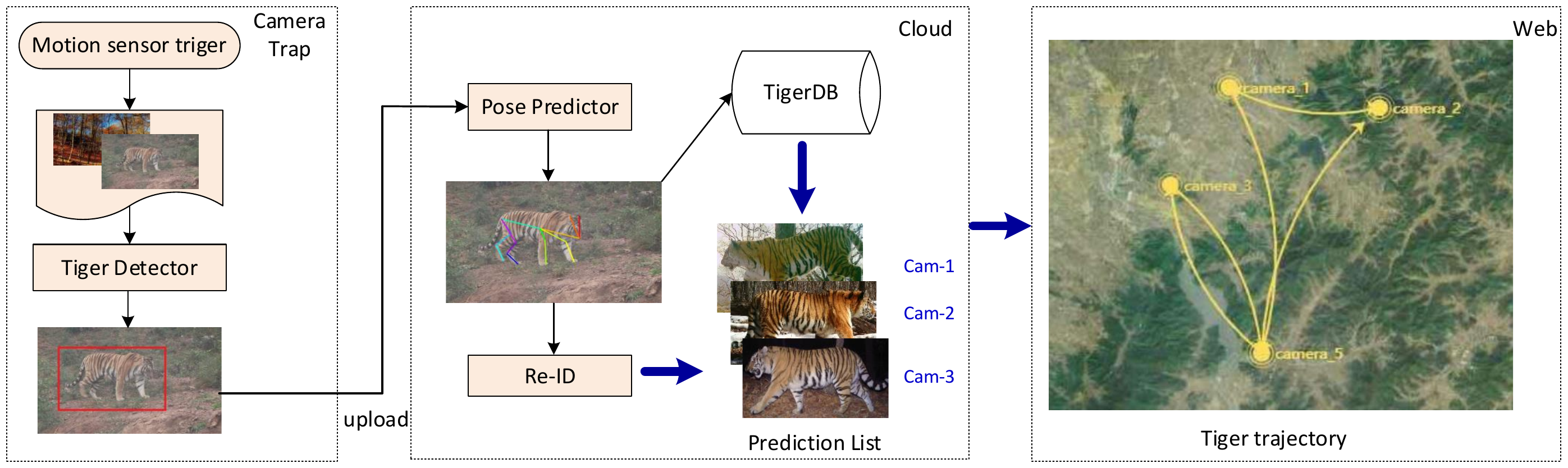}

    \caption{Framework of Amur tiger re-ID system. Motion sensor triggers lightweight detector on smart camera to further filter non-tiger images. Tiger related images are fetched or uploaded to cloud services, which will run pose estimation and re-ID algorithm to produce tiger to camera association. As cameras are discrete placed in a conservation park, we could visualize the tiger moving trajectory on the park map based on camera position information.}
    \vspace{-2ex}
    \label{fig:system}
\end{figure*}

Computer vision techniques are a promising approach to wildlife monitoring, especially with the use of unmanned aerial vehicles or camera traps to collect visual data \cite{hodgson2016}. In particular, re-identification (re-ID) is a core vision method required to obtain accurate population counts and track wildlife trajectory. \autoref{fig:system} illustrates one such system which tracks the movement trajectory of individual Amur tigers through an edge-to-cloud re-identification framework. Amur Tigers are classified as an endangered species, with a remaining population fewer than 600.

However, deployment of such systems is hampered by several challenges. \textit{First}, resource constraints on the edge camera require low-power and accurate tiger detection to trigger the image capture \cite{tinydsod,yolo3,mobilev2} and thus avoid that massive irrelevant image capturing consumes space of storage card and battery life.
\textit{Second}, recent re-ID methods \cite{tripletdefense,Schroff2015FaceNet} typically use pedestrians and cars as target objects, which usually have limited pose variations in a relatively constrained environment. In contrast, wildlife data have a wide range of pose variations due to unrestricted four-limbed movement, complex natural backgrounds, and unconstrained lighting conditions.
\textit{Third}, research into these open challenges is slow due to the lack of datasets and benchmarks beyond object types such as pedestrian and cars \cite{vehicl_dataset,market1501,prw} that have the aforementioned weaknesses. Importantly, most existing datasets lack a systematic benchmark protocol to evaluate end-to-end re-ID performance. This incurs the strong requirement to build a new dataset and benchmark to systematically study wildlife re-identification.

To address the above challenges, we present a novel large-scale dataset named Amur Tiger Re-identification in the Wild (ATRW). Identification of individual Amur Tigers are mostly based on the body stripe patterns \cite{tiger_dataset}, which can be easily deformed due to the posture and movement of the tiger. Manual corrections as in \cite{tiger_dataset} are labor intensive and not suited for wild environments. Automatic correction requires identifying pose keypoints, and potentially additional advanced modeling methods to account for non-rigid body deformation due to tiger movement. Novel methods developed to address these issues can extend to the monitoring of other wildlife species or objects that rely on deformable body patterns for re-identification.

In summary, our major contributions are:
\begin{itemize}
\item We build a new large-scale wildlife dataset ATRW, which contains 92 Amur tiger identities from multiple wild zoos, 8,076 high-resolution video clips from multiple cameras in which tigers are annotated with bounding box positions, pose keypoints and identities on sampled frames.
\item We create a systematic benchmark and comprehensive baseline on ATRW for the full tiger recognition pipeline, including: Amur tiger detection, pose estimation, and re-identification.
\item We propose a novel solution to tiger re-identification based on precise pose parts modeling with deep neural networks to handle the large pose variation of tigers, which demonstrates noticeable performance improvement over traditional re-ID approaches.
\end{itemize}


\section{Related Work}
\begin{table*}[]
    \centering
    \caption{Comparison of animal re-ID datasets. `*' denotes number of video clips, and `-' denotes no public data available. Our dataset is significantly larger, captured in the wild, with rich and dense bounding box annotations, as well as pose keypoint annotations.}
    \label{tab:dataset_comp}
    \begin{tabular}{l |c| c c c c c}
    \hline
    Datasets      &  {\bf ATRW}  &  \cite{tiger_dataset_2,tiger_dataset_1}  &  C-Zoo\cite{czoo_ctai}  &  C-Tai\cite{czoo_ctai}  & TELP\cite{Elephants} &  $\alpha$-whale\cite{whale}  \\
    \hline
    Target        & Amur Tiger & Tiger & Chimpanzees & Chimpanzees & Elephant & Whale\\
    Wild      &$\surd$& $\surd$ & $\times$ &$\times$  &$\times$ & $\surd$\\
    Pose annotation  &$\surd$& $\times$ &$\times$&$\times$  &$\times$ & $\times$\\
    \hline{}
    \#Images or \#Clips   & 8,076$^*$ & \multirow{3}{*}{-}   & \multirow{3}{*}{2,109}  & \multirow{3}{*}{5,078}  & \multirow{3}{*}{2,078} & \multirow{3}{*}{924}\\
    \#BBoxes      & 9,496             &    &   &  &   & \\
    \#BBoxes with ID&  3,649          &    &   &  &   & \\
    \hline{}
    \#identities  &  92   & 298   & 24     & 78    &  276   & 38\\
    \#BBoxes/ID   &  39.7 & -     & 19.9   & 9.7   &  20.5  & 24.3\\
    \hline
    \end{tabular}
\end{table*}

\noindent\textbf{Re-ID datasets.} Several large scale person re-ID datasets have been released in recent years  \cite{CUHK03,DukeMTMC-reID,market1501} to support researches to improve algorithms performance and robustness. Vehicles are also an important object for re-ID due to its wide applications in video surveillance \cite{vehicl_dataset}. Most re-ID datasets only contain cropped images, with the exception of PRW \cite{prw}, which provided raw frames along with annotated bounding box for evaluation of the full re-ID pipeline.

Besides the plethora of person or vehicle re-ID datasets, there are also a few datasets on animal re-ID, which are well summarized in the review \cite{schneider2018past}, including primates \cite{Primates,czoo_ctai},
tigers \cite{tiger_dataset_2,tiger_dataset_1}, elephants \cite{Elephants}, and whales \cite{whale}. However, these animal re-ID datasets have various weaknesses, such as small data sizes, limited annotations, and captured in non-wild settings. Because of these limitations, these datasets are not widely used in re-ID research. Our contributed dataset, ATRW, fills this gap to provide a large-scale, well-annotated, full pipeline re-ID dataset, to challenge existing approaches. For a comparison with existing animal re-ID datasets, see \autoref{tab:dataset_comp}.
\textit{People may concern that the identity number in our ATRW is relatively small (92). However, we should emphasize that the total number of wild Amur tigers is less than 600, so that our collected indentity number already reflects practical usage requirement. And this is common for the re-identification of all the endangered species}.

\noindent The term of \textbf{Re-identification} was first proposed in 2005 \cite{firstreid} to support multi-camera person tracking with explicit ``re-identification'' based on appearance features.
With large scale datasets, deep learning approaches become dominant in this field. Many approaches learn identity-related representations for re-ID purposes \cite{reid_class,reid_attr}.
Others formulate re-ID as a ranking problem, and feed a pair of images into a convolutional neural network to learn a ranking function \cite{Ahmed2015An,CUHK03,Yi2014Deep}.
Another type of approaches formulates re-ID as a metric learning problem, and combine CNNs with novel loss functions to learn similarity metrics \cite{tripletbeyond,Schroff2015FaceNet}. Local modeling with fusion of global and local representations is also used to enhance re-ID  \cite{wei2017glad,aligned_reid,zheng2017pose}.

Successful deployment of re-ID in the wild also requires object detection and pose estimation methods to normalize the image for accurate matching. \textbf{Efficient object detection} on the edge client is an active area of research. Typical methods include MobileNet-SSD \cite{mobilev1,mobilev2}, SqueezeDet \cite{Wu2017SqueezeDet}, and TinyDSOD \cite{tinydsod}.
\textbf{Pose estimation} is useful for precise target modeling especially in datasets with rich variations in pose. Human pose estimation is relatively well studied with datasets such as COCO \cite{MSCOCO} and MPII\cite{MPII}. To our knowledge, there are no re-ID datasets that include annotated ground truth for pose estimation, and also no datasets that permit studying how pose estimation impacts the re-ID performance. Our ATRW dataset will provide ground truth annotations for tiger pose, as well as studies of the impact of generic pose estimation methods such as OpenPose \cite{openpose}, AlphaPose \cite{alphapose} and HRNet \cite{hrnet}.


\vspace{-1ex}
\section{The ATRW Dataset}
\subsection{Annotation Description}
The Amur tiger (also known as Siberian tiger, Northeast-China tiger) is a tiger population in the far east region (particularly the Russian Far East and Northeast China), which currently has less than 600 wild individuals in the world. Capturing enough image data for free-roaming Amur tigers is infeasible as these tigers have an activity range over hundreds of kilometers. Instead, we capture Amur tigers from multiple large wild zoos with the help of the World Wildlife Foundation (WWF). Images are collected in unconstrained settings with time-synchronized surveillance cameras and tripod fixed SLR cameras. In total, we captured 8,076 high resolution (1920$\times$1080) video clips with at least one Amur tiger, which are further uniformly sampled (one out of ten) into frames. Some frames are discarded due to motion artifacts, lack of tigers, or other noise.

The annotation contains three steps.
\textit{First}, we annotate a bounding box for each tiger as well as the view orientation of the tiger in the image (frontal, left, right, back).
\textit{Second}, professionals determine the tiger identity based on temporal and appearance cues.
Tigers from different zoos are physically separated during annotation, so that the id annotation is fairly accurate.
If the tiger cannot be clearly identified, an unknown id will be assigned to that tiger, which will not be used in the re-ID procedure.
\textit{Third}, as the tiger movement causes large pose variations, we further annotated skeleton keypoints for each tiger for downstream pose normalization or precise pose modeling.
Details of the keypoint definition will be discussed in the following section.

Similar to many other re-ID datasets, we also provide a cropped dataset uncoupled from the wild environment for isolated tests of re-ID algorithms.
The annotated images are cropped according to the bounding boxes, and renamed with camera id, shot id, frame number and entity id.
Here we use entity to refer to a combination of the tiger identity and its side information.
Based on professional consultations \cite{tiger_dataset}, the tiger stripe pattern is the most informative marker of tiger identity. Since the left and right side of Amur Tigers have different stripe patterns, and it is rare to capture both sides of the tiger in the wild environment, we treat different sides of the same tiger as a different entity.

The full pipeline of tiger re-identification contains three modules as shown in \autoref{fig:system}. We describe data information of each module in detail below.
\vspace{-2.5ex}
\subsubsection*{Detection Data}
\begin{figure}[]
    \centering
    \subfigure[width distribution]{
    \includegraphics[width=0.475\linewidth]{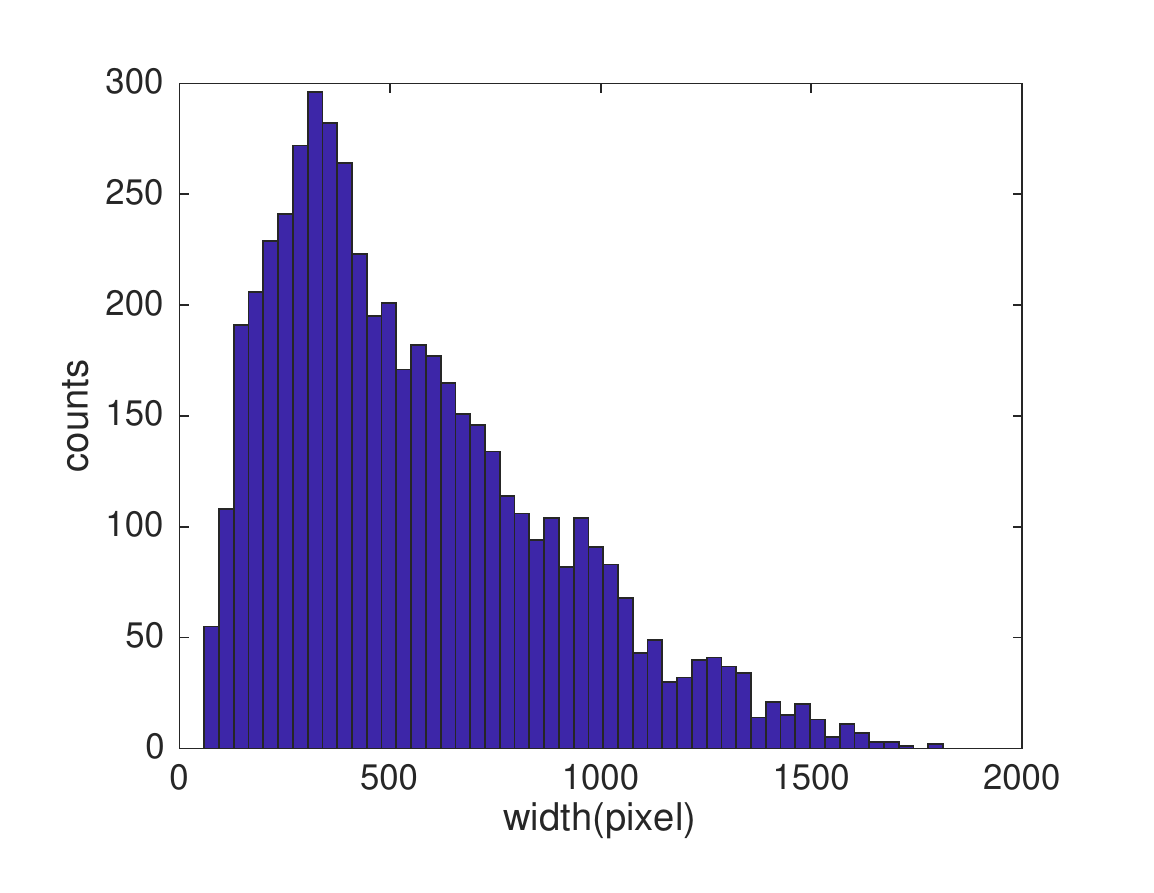}
    }
    \subfigure[aspect ratio distribution]{
    \includegraphics[width=0.475\linewidth]{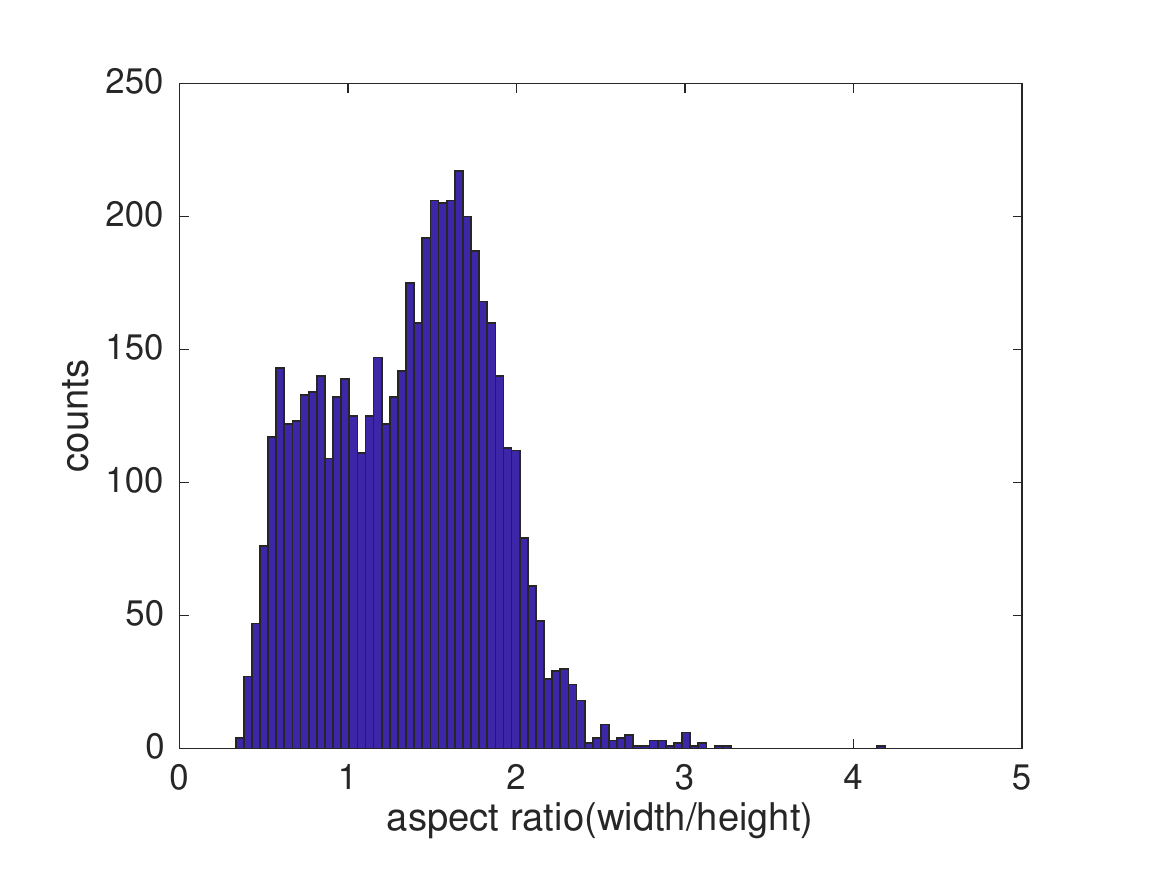}
    }
    \vspace{-3ex}
    \caption{Distribution of width and aspect ratio (width/height) of bounding boxes in the dataset.}
    \label{fig:dis:detection}
\end{figure}

The object detection module allows the system to select only frames that include an Amur tiger, thus reducing storage, power, and networking consumption. Our tiger detection dataset includes 4,434 images with 9,496 bounding boxes. Some of the bounding boxes may have the `unknown' tiger identity, as described above, but the annotations can still be used for training and testing object detector models. We provide annotations in the same format as that of PASCAL VOC \cite{VOC}. See \autoref{fig:exp_det} for some sample bounding boxes. The distribution of bounding box width and aspect ratio are shown in \autoref{fig:dis:detection}.

\subsubsection*{Pose Keypoint Data}
\begin{figure}[h]
    \centering
    \includegraphics[width=0.7\linewidth]{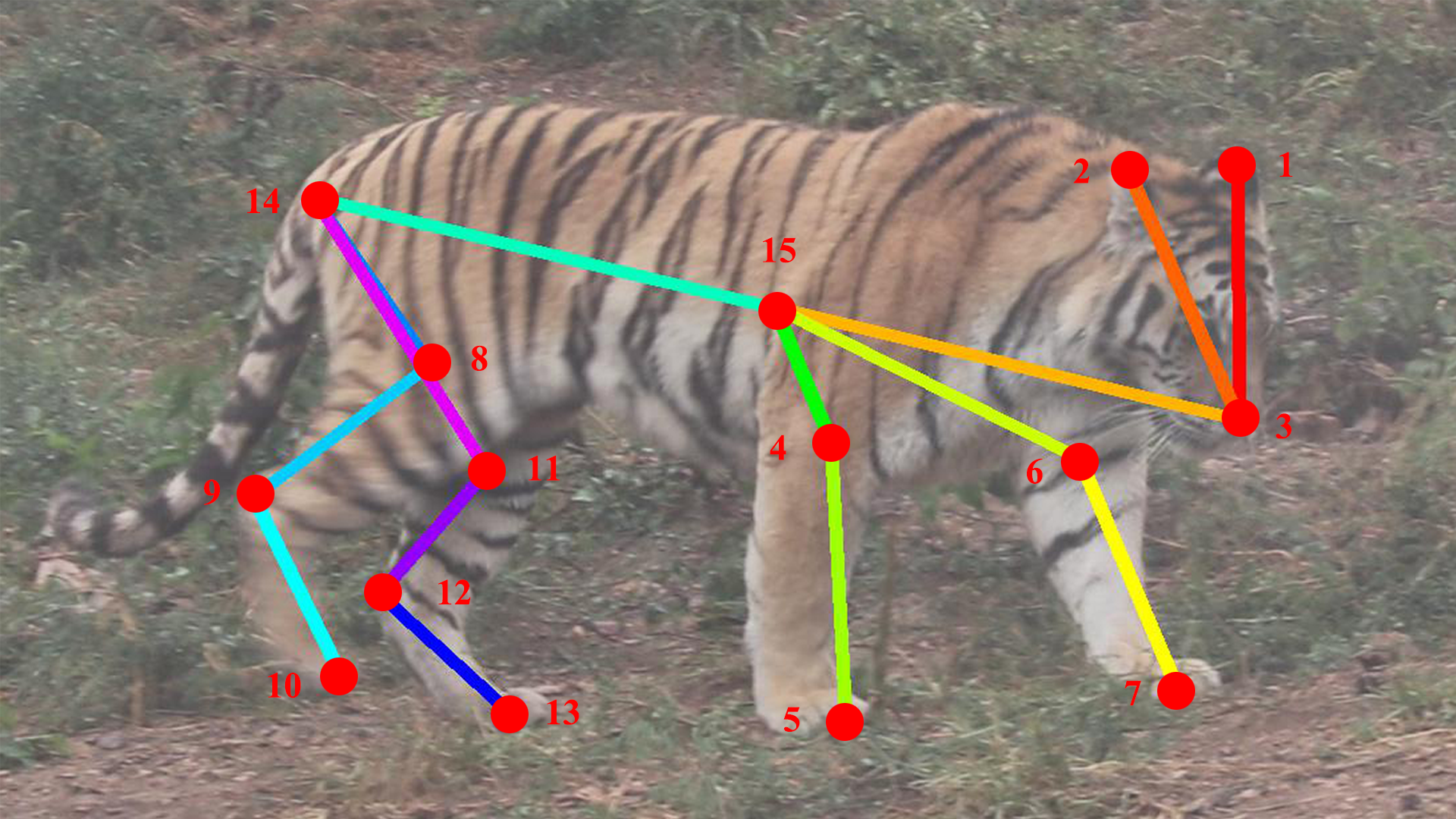}
    \caption{Definition of tiger keypoints.}
    \label{fig:pose_def}
\end{figure}

\begin{figure*}[]
    \begin{minipage}{1\linewidth}
    \centering
        \includegraphics[width=0.22\linewidth]{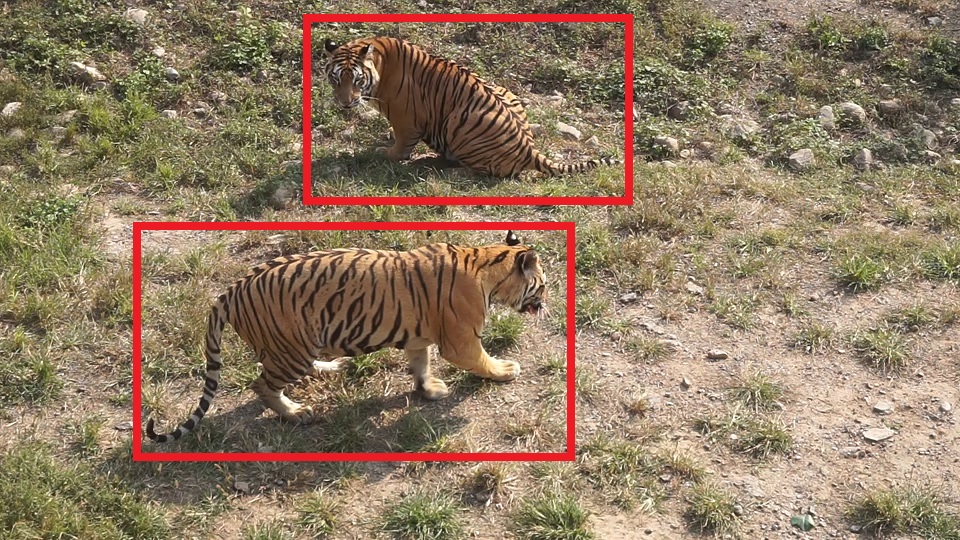}
        \hspace{0.0075\linewidth}
        \includegraphics[width=0.22\linewidth]{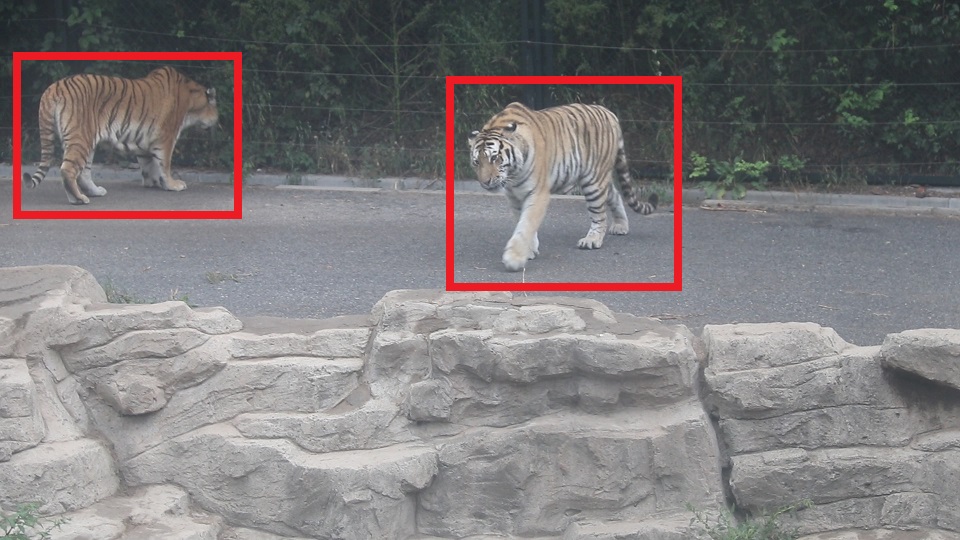}
        \hspace{0.0075\linewidth}
        \includegraphics[width=0.22\linewidth]{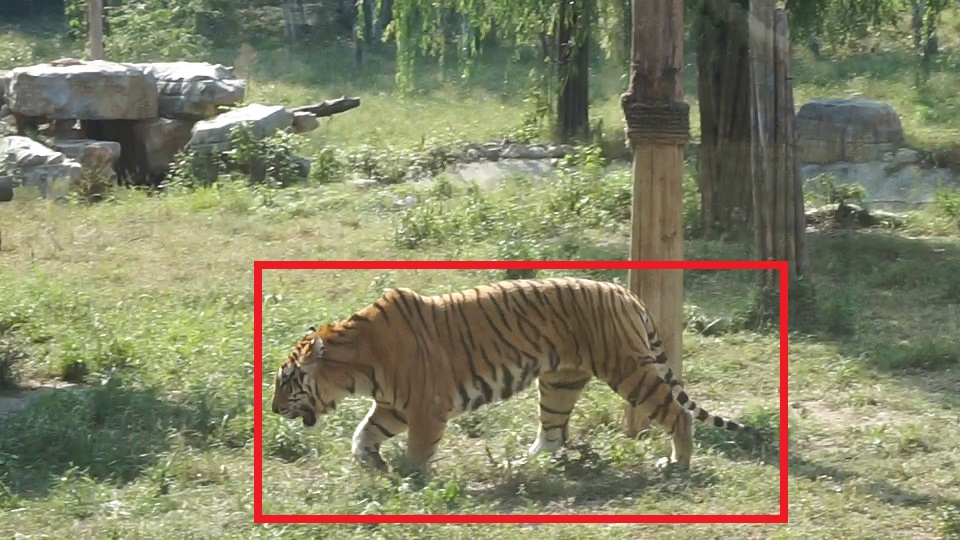}
        \hspace{0.0075\linewidth}
        \includegraphics[width=0.22\linewidth]{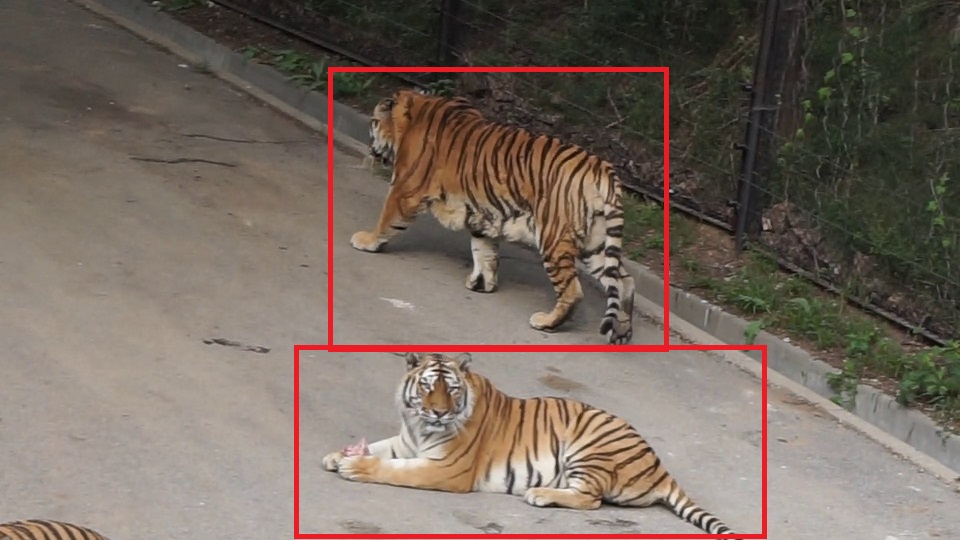}
        \caption{Example images of detection subset.}\label{fig:exp_det}
    \end{minipage}
    \begin{minipage}{1\linewidth}
    \centering
        \includegraphics[width=0.22\linewidth]{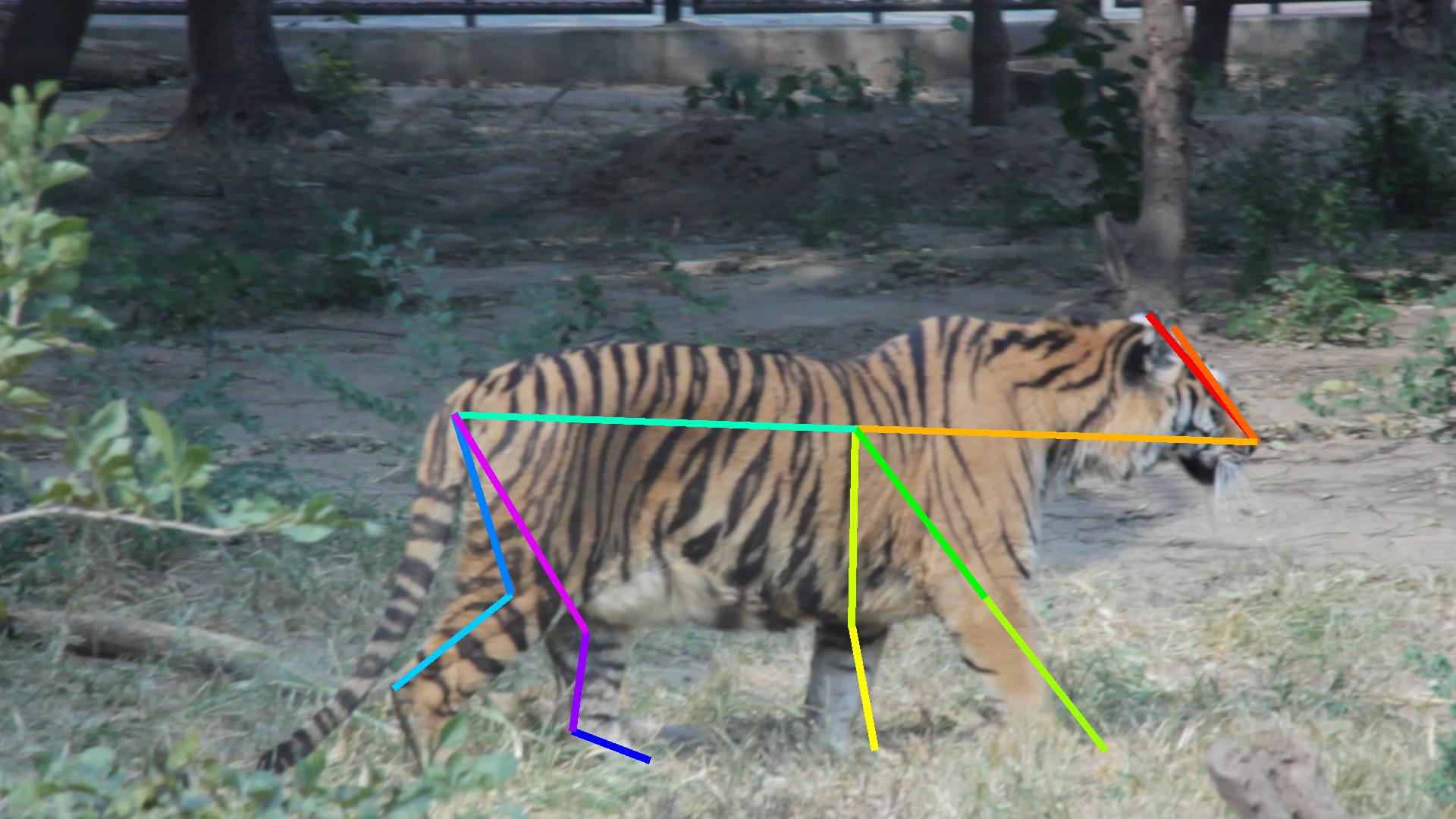}
        \hspace{0.0075\linewidth}
        \includegraphics[width=0.22\linewidth]{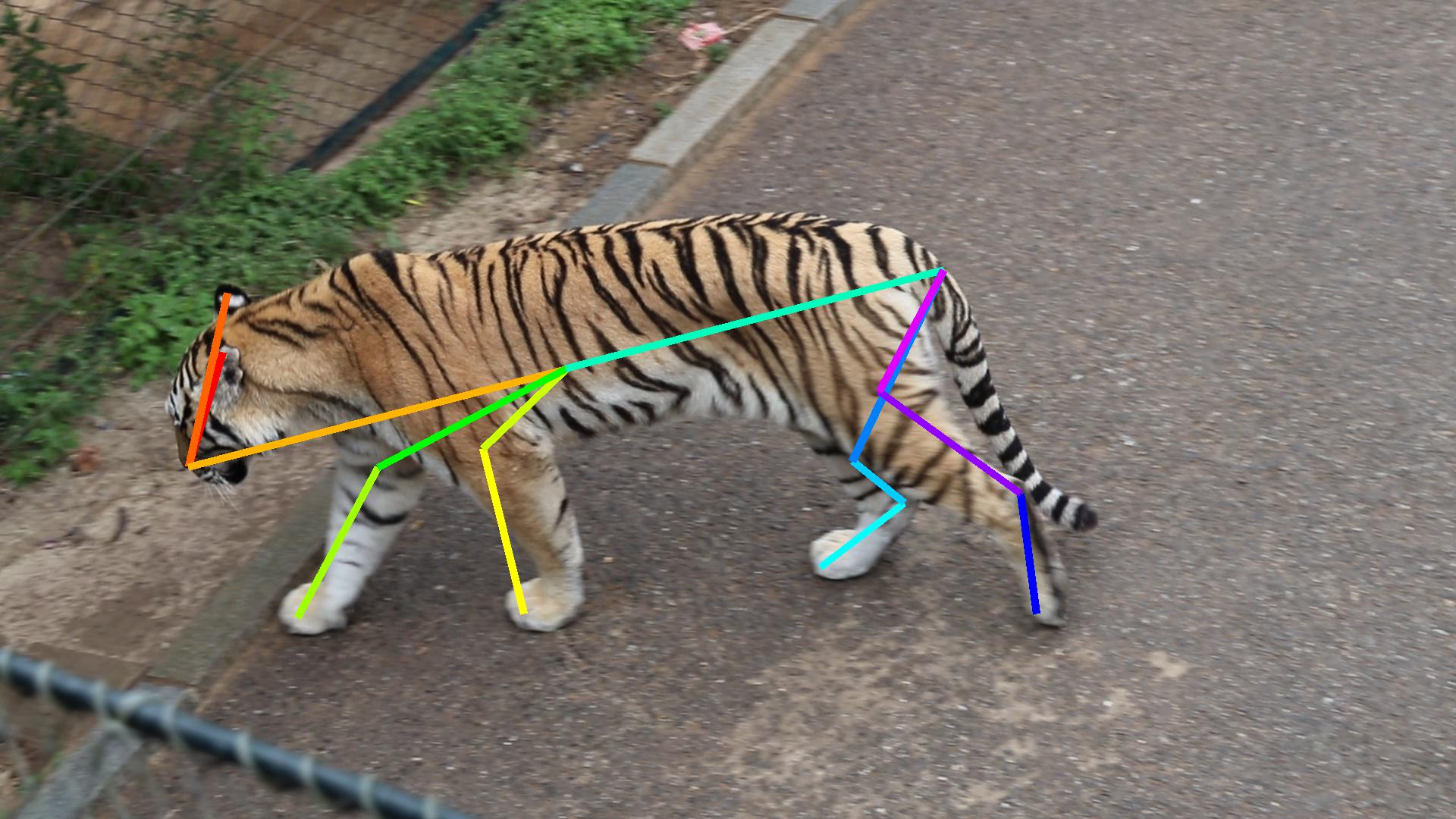}
        \hspace{0.0075\linewidth}
        \includegraphics[width=0.22\linewidth]{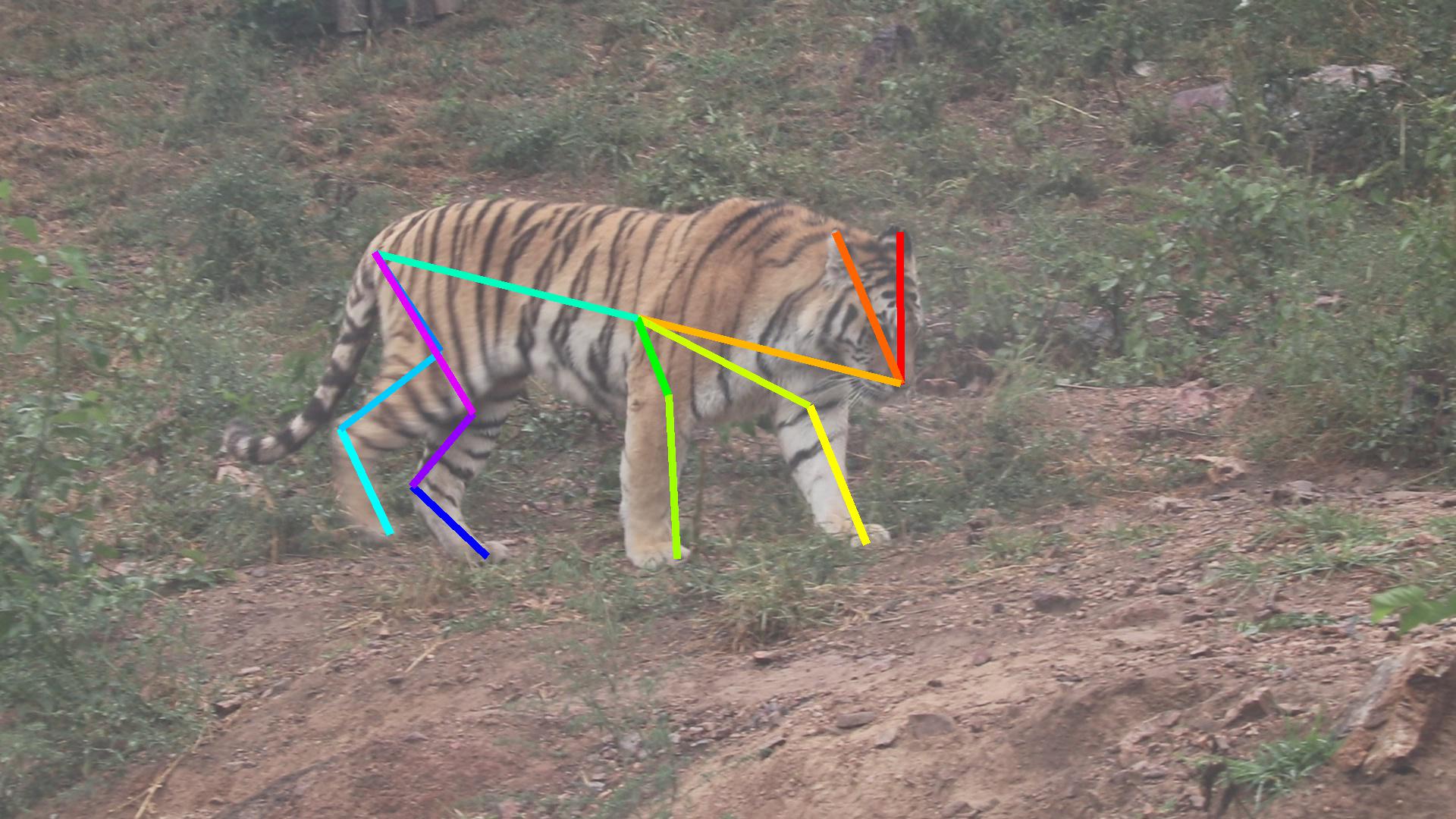}
        \hspace{0.0075\linewidth}
        \includegraphics[width=0.22\linewidth]{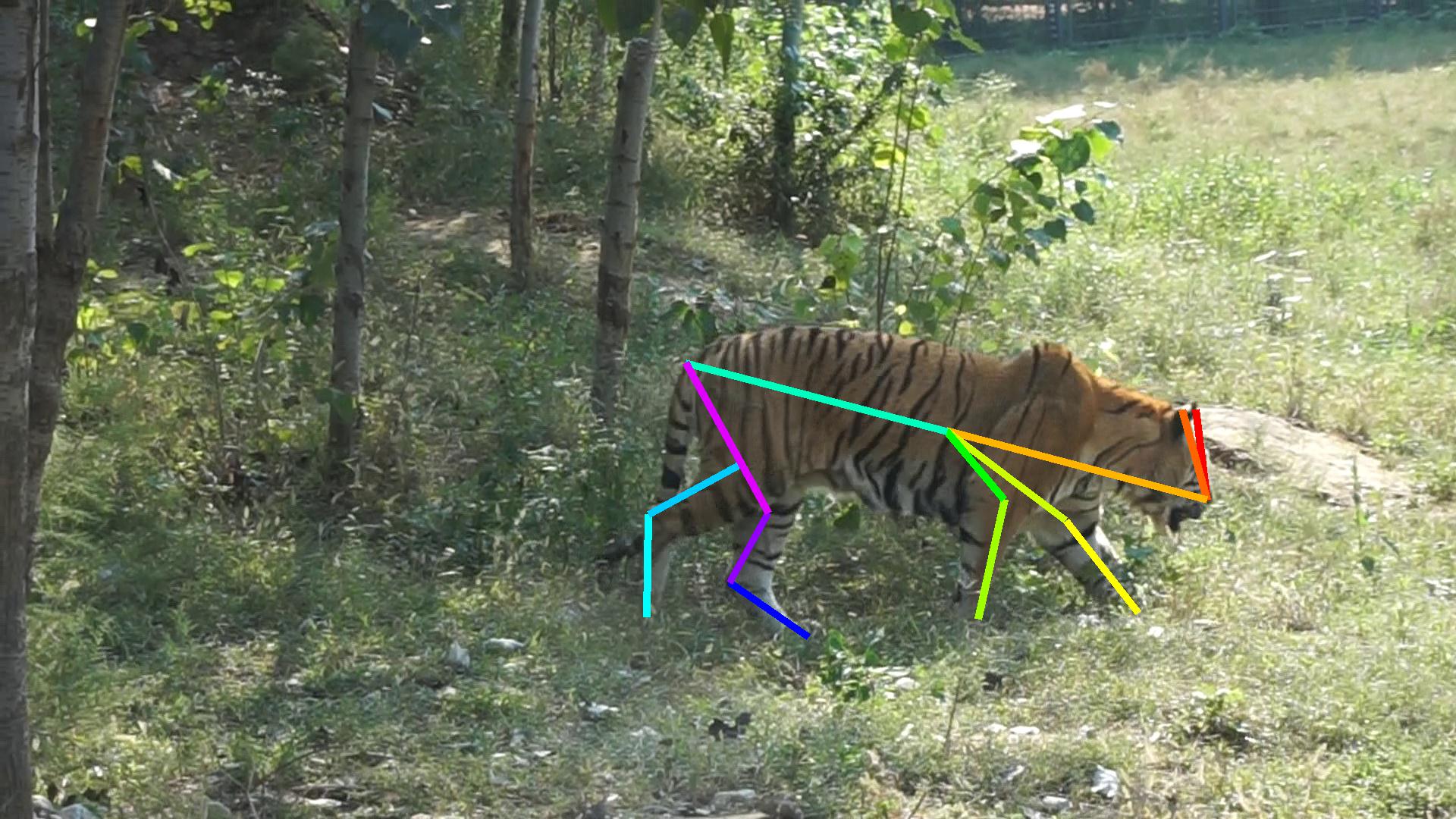}
        \caption{Example images of pose subset.}\label{fig:exp_pose}\label{fig:dataset_sample}
    \end{minipage}
\end{figure*}

This module performs pose estimation to locate tiger skeleton keypoints, which are important to the re-ID task due to the significant non-rigid movement of the tigers. The extracted pose information could be used to align and normalize the tiger, or provide precise modeling of the tiger to improve the accuracy of re-ID algorithms.
We defined a set of tiger skeleton keypoints (\autoref{tab:key_point_def}), as illustrated in \autoref{fig:pose_def} and \autoref{fig:exp_pose}.
If there are more than two annotations per tiger, the keypoint positions are averaged when the annotations are not too far apart. If there are significant differences, we manually validate the keypoints. The annotation is given in the COCO format.

\begin{table}[h]
    \centering
    \caption{Definition of key-points in our dataset}
    \scriptsize
    \begin{tabular}{c|l||c|l}
    \hline
    key-point & definition & key-point & definition\\
    \hline
    1  & left ear        &  9  &  right knee\\
    2  &  right ear      & 10 &  right back paw\\
    3  &  nose           & 11 &  left hip\\
    4  &  right shoulder & 12 &  left knee\\
    5  &  right front paw& 13 &  left back paw\\
    6  &  left shoulder  & 14 &  root of tail\\
    7  &  left front paw & 15 & center, mid point of 3 \& 14\\
    8  &  right hip &&\\
    \hline
    \end{tabular}
    \label{tab:key_point_def}
    \vspace{-3ex}
\end{table}

\subsubsection*{Re-ID Data}
Stripe information is used for tiger re-ID based on suggestions from professionals, so that we focus on images of the left and right side of tiger body.
Each side is viewed as different entity of tiger as mentioned before.

We define two settings to evaluate re-ID algorithms.
First, in `plain re-ID' setting, both the query tigers and database tigers are cropped and normalized with manually annotated bounding boxes and poses. Second, `wild re-ID' requires automatic tiger detection and pose estimation to provide tiger normalization for the following re-ID procedure.

\begin{figure}[h]
\centering
\small
\includegraphics[width=0.6\linewidth]  {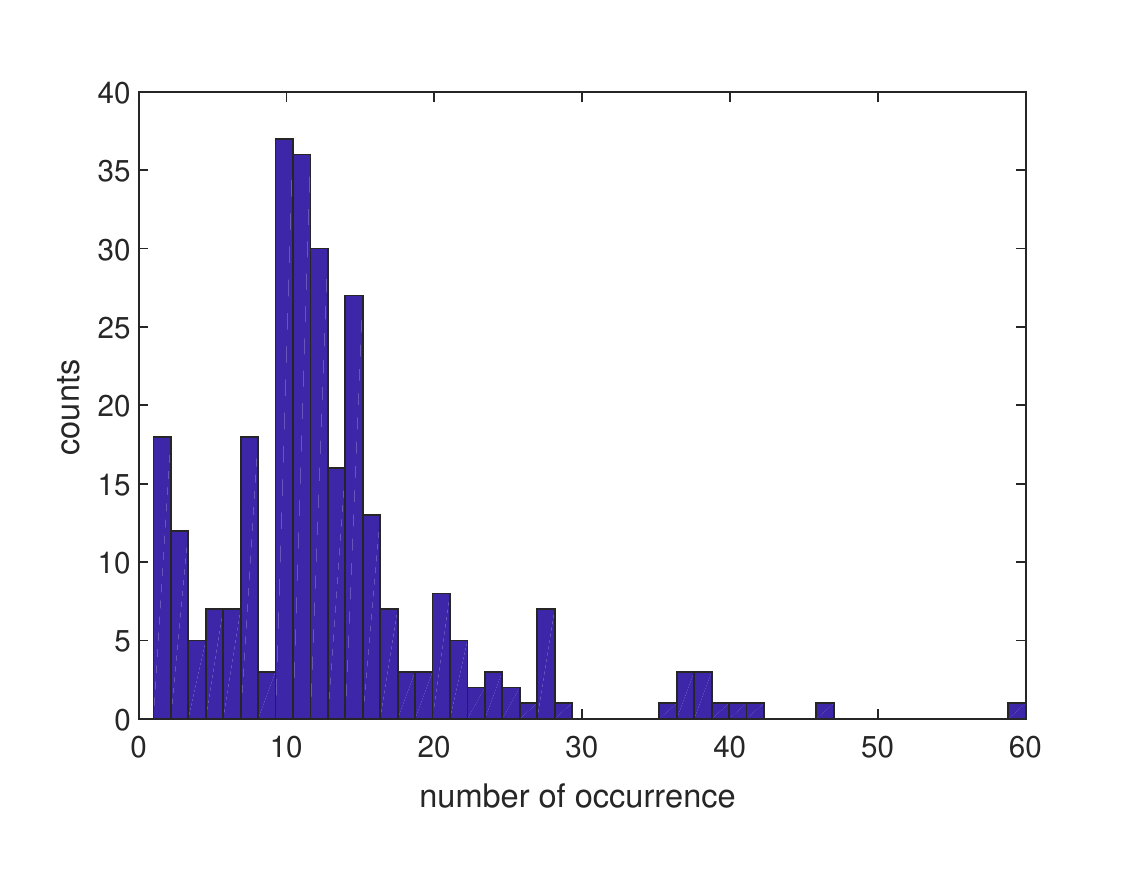}
\vspace{-1ex}
\caption{Distribution of occurrence times.}\label{fig:stat:bbox_per_entity}
\end{figure}

The re-ID dataset contains 182 entities of 92 tigers, with a total of 3,649 bounding boxes. Most of the entities appear at least 10 times in the subset, as shown in the histogram of the number of occurrences in \autoref{fig:stat:bbox_per_entity}.
Unlike the popular person re-ID dataset Market-1501~\cite{zheng2015scalable}, not all entities appear cross camera due to capturing restrictions in some wild zoos. In this case, we ensure that significantly different frames from the same camera are selected into our dataset.
\autoref{Sing_Cross} lists the detailed data distribution among cross-camera and single-camera, showing that each entity has an average of 28.3 bboxes. As aforementioned, entity is the basic unit of re-ID, which is a single side of a tiger.

\begin{table}[h]
\centering
\caption{Individual distribution: single-camera vs cross-camera}\label{Sing_Cross}
\small
\begin{tabular}{l|c c c}
\hline
 & \#Entity & \#Tiger  &\#BBox\\
\hline
Single-Cam &132 & 53& 1927\\
Cross-Cam  & 50 & 39& 1722\\
Total      &182 & 92& 3649\\
\hline
\end{tabular}
\end{table}

\vspace{-2ex}
\subsection{Evaluation Protocol}\label{sec:eval}

\subsubsection*{The detection task}
To conform with the following wild re-ID task and avoid possible ground truth bounding box annotation leakage, we split 1,651 images (about 37\%) as testing set for the detection task, which contains images from all tiger entities in the testing set of the wild re-ID task, while keep the other images as the training set, in which a portion of training subset (10\%) are used for validation purposes.
Similar to the COCO object detection challenge \cite{MSCOCO}, we use the average precision (AP) at different Intersection-over-Union (IoU) thresholds to evaluate the performance.
As detectors will be deployed on resource-constrained edge devices, we combine mAP and FLOPs together to propose Performance per FLOPs (PPF) to evaluate detectors in comparison to a baseline with 0.43 mAP and 1 Billion FLOPs as below:
\begin{equation}
\small
\label{equ:ppf}
PPF=\frac{mAP-0.43}{BFLOPs}.
\end{equation}

\subsubsection*{The pose estimation task}
The pose estimation task also randomly splits the dataset into training and testing subset, containing 80\% and 20\% bboxes with keypoint annotations respectively.
The evaluation metric is AP measured by Object Keypoint Similarity (OKS) as defined in COCO \cite{MSCOCO}.
\autoref{tab:stat:keypoint} lists the annotation statistics for each keypoint due to inhomogeneous annotations from different annotators, where $\sigma^2$ denotes the variance of keypoint position normalized by object scale.
Formally, $\sigma_i^2=E[d_i^2/s^2]$, $d_i$ represents the deviation of the $i$-th keypoint among different annotators, and $s$ represents the scale of the object. Then OKS is calculated as
\begin{equation}
\small
\label{equ:oks}
OKS=\frac{\sum_{i}[\delta(v_i>0)\exp(-\frac{d_i^2}{2s^2k_i^2})]}{\sum_{i}[\delta(v_i>0)]},
\end{equation}
where $k_i=2\sigma_i$ and $v_i>0$ if the $i$-th keypoint is visible.

\begin{table}[h]
    \centering
    \caption{Variance of keypoints annotations.}\label{tab:stat:keypoint}
    \scriptsize
    \begin{tabular}{c|r||c|r||c|r}
    \hline
    keypoint & $\sigma^2(10^{-4})$ & keypoint & $\sigma^2(10^{-4})$& keypoint & $\sigma^2(10^{-4})$\\
    \hline
    1  & 7.7  & 6  & 6.9  & 11 & 11.1\\
    2  & 67.7 & 7  & 41.7 & 12 & 29.9\\
    3  & 69.0 & 8  & 9.1  & 13 &  6.9\\
    4  & 4.1  & 9  & 19.4 & 14 & 46.7\\
    5  & 51.3 & 10 & 10.0 & 15 & 29.0\\
    \hline
    \end{tabular}
\end{table}

\begin{table}[h]
    \centering
    \caption{Statistics of training set and test set for re-ID task}\label{tab:stat:train_test}
    \small
    \begin{tabular}{c|c c c}
    \hline
    Datasets & \#images & \#entities & \#tigers \\
    \hline
    Train & 1887 & 107 & 75  \\
    Test  & 1762 & 75  & 58 \\
    \hline
    \end{tabular}
\end{table}
\begin{table}[h]
    \centering
    \caption{Statistics of query from cross-camera and single-camera cases in the re-ID task.}\label{tab:stat:query}
    \small
    \begin{tabular}{c|c c c}
    \hline
    Queries     & \#images & \#entities & \#tigers  \\
    \hline
    Single-Cam. & 701  & 47 & 42 \\
    Cross-Cam.  & 1061 & 28 & 20 \\
    \hline
    \end{tabular}
\end{table}
\subsubsection*{The re-ID task}
As described previously, we have defined two tracks (plain re-ID and wild re-ID) based on whether
images are cropped manually or automatically.
Similar to Market-1501~\cite{market1501}, we choose the mean average precision (mAP) as the primary metric to evaluate the performance and top-$k$ accuracy as the secondary metric. For each query, the Re-ID algorithm produces a list of predictions from which we measure the AP.
The mean value of AP scores over all queries is the mAP, which is based on the evaluation code from Market-1501 \cite{market1501}, but with following difference.
In Market-1501, if the algorithm returns a result image that is from the same cameras as the query image, a false positive is counted. In our dataset, we sometimes have tiger identities that are only captured from a single camera. Therefore, we modified the handling of returning images from the same camera slightly differently.
We exclude temporal adjacent images within 1 second (forward and backward) to the query image from the query results for computing the AP score.
To be more precise, we separate each query image into two categories: `single camera', where the identity only appears in one camera, or `cross-camera', where the target appears in multiple cameras. We report performance for both cases.

We then constructed the training and testing sets with the following procedure. First, we randomly choose 60\% entities from the single-camera category and 40\% entities from cross-camera category. Collectively, this formed the training set. The remaining images comprise the test set. Each image in testing set will be queried once with respect to the whole testing set. Metrics are calculated separately for single-camera and cross-camera part in testing set. Detailed statistics about dataset split can be found in \autoref{tab:stat:train_test} and \autoref{tab:stat:query}.

As aforementioned, the re-ID task consists of both the `plain re-ID' case and the `wild re-ID' case.
The plain case uses manual tiger bounding-box and pose annotations for the re-ID purpose, while the wild case
aims to evaluate full-pipeline performance based on automatic tiger detection and pose estimation results.
More strictly, bounding-boxes of both the gallery and query are generated by detection module.
For those query tigers not found by the detectors, the corresponding query AP is counted as 0.
\vspace{-1ex}
\section{Baseline Methods}\label{sec:baseline}
Here we describe the baseline methods we used for re-ID, especially including our innovative pose part-based CNN modeling framework. We use common baselines for object detection and pose estimation, which we will introduce with the results.
\vspace{-1ex}
\subsection{Classification based Baseline} 
\begin{figure*}[]
    \centering
\begin{minipage}{.42\linewidth}
    \centering
    \includegraphics[width=0.9\linewidth]{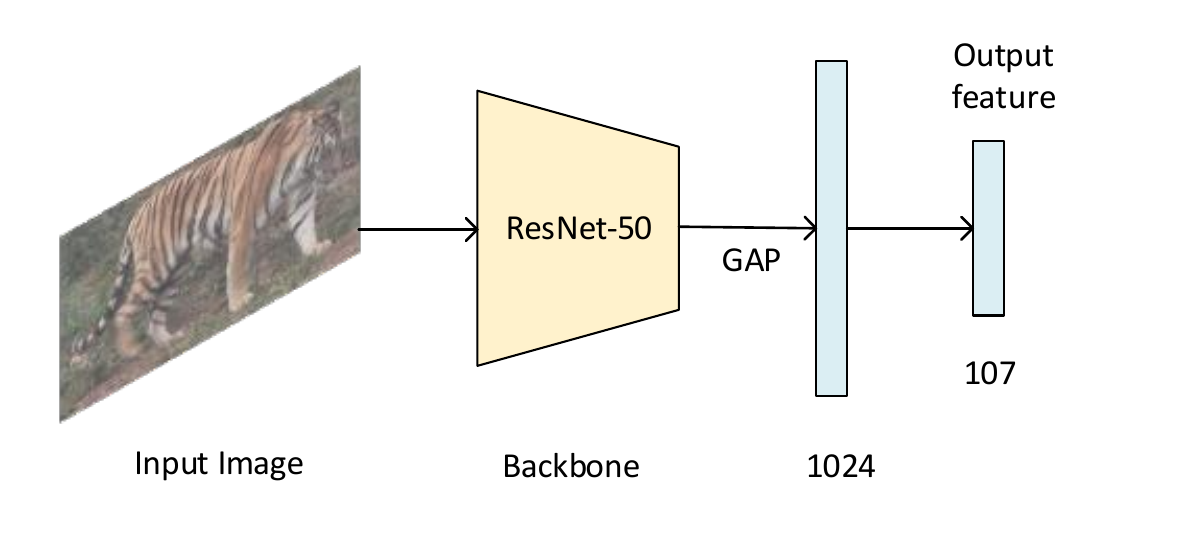}
    \caption{Architecture for classification/triplet based baseline.}\label{fig:net_arch_base}
\end{minipage}
\hspace{1ex}
\begin{minipage}{.485\linewidth}
    \centering
    \includegraphics[width=0.99\linewidth]{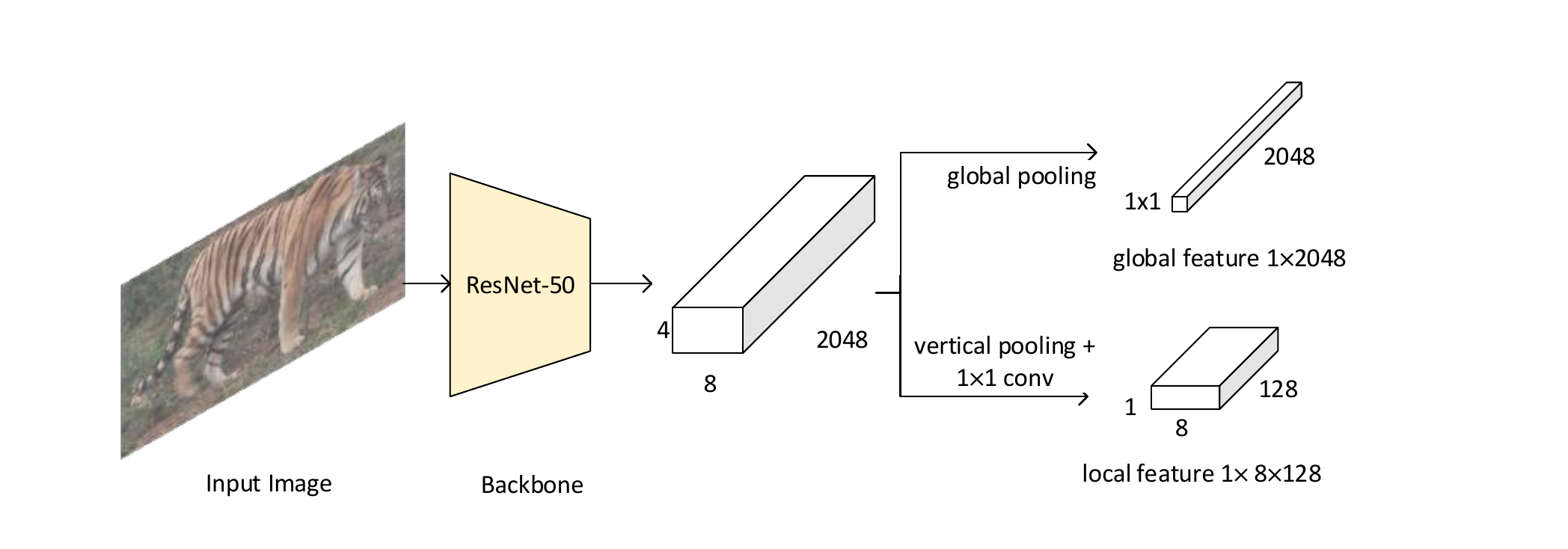}
    \caption{Architecture for Aligned-ReID baseline.}\label{fig:net_arch_aligned}
\end{minipage}
\end{figure*}

Large-scale trained classification networks are believed to produce a rich representation that could be generalized to new tasks, particularly under the same dataset. Hence, we use a classification based re-id method as proposed in \cite{reid_class} as a baseline. That model uses an ImageNet pre-trained ResNet-50~\cite{resnet} backbone, followed by two fully-connected (FC) layers with 1024 and 107 neurons respectively (\autoref{fig:net_arch_base}).
The input image resolution is $256\times 128$ to accommodate the horizontal aspect ratio of tigers.
The output feature length of the network ($n=107$) is set to the number of entities in the training set. To avoid overfitting, the ResNet-50 backbone parameters are frozen, and we only train the added classification layers.
As classification tasks usually use a cross entropy (CE) loss, we refer this baseline as CE.
\vspace{-1ex}
\subsection{Triplet Loss Baseline} 
Metric learning is also widely used for re-ID, including methods such as Triplet loss\cite{Schroff2015FaceNet}, Quadruplet loss\cite{tripletbeyond}, or TriHard loss \cite{tripletdefense}. Triplet loss aims to pull semantically similar points on the data manifold close in the embedding space and push dissimilar points farther apart.
In this study, we choose TriHard, a triplet loss variant with batch hard mining, as one of our baseline method. TriHard loss examines the hardest pairs between positive pairs and negative pairs in a mini-batch.
Formally, the triplet loss is
\begin{equation}
\small
    \mathcal{L}_{tri}=[\|f(x_a)-f(x_p)\|_2-\|f(x_a)-f(x_n)\|_2+m]_+,
\end{equation}
    where $x_a$, $x_p$, $x_n$ represents the anchor, positive and negative input respectively, $f(\cdot)$ represents the network, $m$ represents a margin. And the TriHard loss is defined as
\begin{align}\label{eq:loss_bh}
\small
\mathcal{L}_{TH}(X) = \sum\limits_{i=1}^{P} \sum\limits_{a=1}^{K}
    \Big[m & + \hspace{-5pt} \max\limits_{p=1 \dots K} \hspace*{-1pt} \left\|f(x^i_a)-f(x^i_p)\right\|_2 \\
           & - \hspace{-5pt} \min\limits_{\substack{j=1 \dots P \\ n=1 \dots K \\ j \neq i}} \hspace*{-1pt} \left\|f(x^i_a)-f(x^j_n)\right\|_2 \Big]_+,\nonumber
\end{align}
where $X$ is a mini-batch of $PK$ samples, which has $P$ identities and $K$ samples from each identity. The max-term represents the farthest positive pairs and the min-term represents the nearest negative pairs.
For simplicity and fair comparison, we used the same network architecture as used in the classification baseline as shown in  \autoref{fig:net_arch_base}.

\begin{figure}[]
    \centering
    \includegraphics[width=0.6\linewidth]{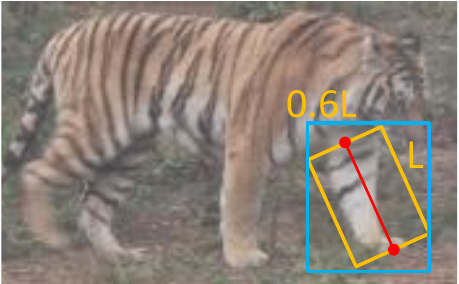}
    \caption{Extract AABB from skeleton. Two red points are keypoints defining a part, while the orange rectangle is the corresponding bounding box with height of L and width of 0.6L. The blue rectangle is AABB which circumscribes the orange bbox.}
    \label{fig:aabb}
\end{figure}

\subsection{Aligned re-ID Baseline}
The above methods use the global representation, which may not be effective for targets with large pose variations. In the aligned re-ID method \cite{aligned_reid}, the authors propose a local distance concept to enhance the feature representation. They use dynamic programming to compute the shortest mapping path between two images, and define local distance as the length of shortest path. The final distance between two images is the sum of global distance and this local distance.
In our implementation, we only introduce the metric loss to combine local distance and global distance, without using other tricks in the original Aligned Re-ID work \cite{aligned_reid}.

Due to the addition of local distance, the network architecture is changed as shown in \autoref{fig:net_arch_aligned}.
For the last pooling layer, we pool along each column to obtain local features $F(x)=\{f_1(x),\dots,f_W(x)\}$ for input image $x$, where $W$ is the pooling feature map width ($=8$ as shown in Figure \ref{fig:net_arch_aligned}). We define some intermediate variables as
\begin{equation}
\small
    d_{i,j}(x, y)=\frac{\exp(\|f_i(x)-f_j(y)\|_2)-1}{\exp(\|f_i(x)-f_j(y)\|_2)+1}  \quad i,j\in 1,2,3\dots,W,
\end{equation}
\begin{equation}
\small
    S_{i,j}(x,y)=\left\{\begin{matrix}
    d_{i,j}           & i=j=1\\
    S_{i-1,j}+d_{i,j} & i\neq 1,j=1\\
    S_{i,j-1}+d_{i,j} & i=1,j \neq 1\\
    \min(S_{i-1,j},S_{i,j-1})+d_{i,j} & i,j \neq 1,
    \end{matrix}\right.
\end{equation}
and the local distance as
\begin{equation}
\small
    D_{local}(x_1,x_2)=S_{W,W}(x_1, x_2).
\end{equation}

A local distance based loss is defined as
\begin{equation}
\small
    \mathcal{L}_{local}=[D_{local}(x_a,x_p)-D_{local}(x_a,x_n)+m]_+,
\end{equation}
where $x_a$, $x_p$, $x_n$ are examined by batch hard mining in TriHard using global distance.
The total loss $\mathcal{L}=\mathcal{L}_{TH}+\mathcal{L}_{local}$ is used for this baseline.

\subsection{Pose Part based Model}
\begin{figure*}[]
    \centering
    \includegraphics[width=0.8\linewidth]{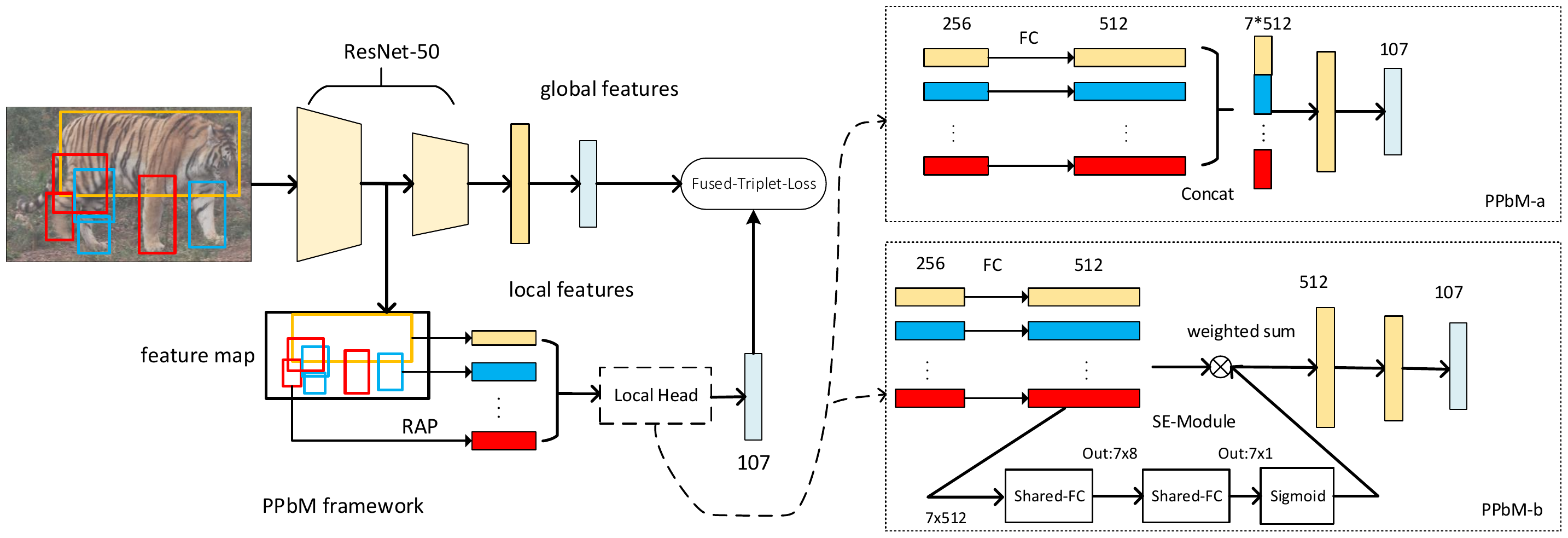}
    \caption{Architecture of our proposed pose part based model (PPbM). Left is the overall framework. Right are two ``local-head'' structures with PPbM-a at top-right, PPbM-b at bottom-right.}
    \label{fig:star_model}
\end{figure*}

While the Aligned-ReID demonstrates strong performances for pedestrian re-ID, the method performs worse than the triplet-loss based baseline for tiger re-ID. Please see experimental section for detailed comparison.
Because tigers have much larger pose variation due to non-rigid motion, the local feature representations created by pooling do not provide an invariant representation or precise modeling of the tiger body.
Part-based models have shown great success for objects that are composed of deformable parts in tasks such as object detection \cite{dpm} or fine-grained object recognition~\cite{zhang2014part}.
This kind of methods represent local parts by a rectangular patch, and adopts structured SVM to learn part structures.
Recently, pose keypoint estimation techniques such as OpenPose \cite{openpose} or AlphaPose \cite{alphapose} provide even precise
body parts and skeletons modeling, which offers new opportunities for part-based model.

We propose pose part based model (PPbM) for tiger re-ID, which seamlessly integrates the result of pose keypoint estimation into deep neural networks.
\autoref{fig:star_model} illustrates the network architecture.
We characterize a tiger with a 7-part star model, including trunk, left and right of front legs, hind thighs, and hind shanks.
For each part, we compute Axes Aligned Bounding Box (AABB) according to the pose skeleton, as shown in \autoref{fig:aabb}. \textit{In fact, non-AABB region representation is more accurate, but less efficient to compute on feature-map during training phase, so that we have to resort to the AABB approximation}.
For each AABB area, we apply the ResNet-50 backbone, and extract the local feature representation with regional average pooling (RAP) on the \textit{res3d} feature map. We use the intermediate layer \textit{res3d} instead of the final residual layer because relative higher feature-map resolution (32$\times$ 16 vs 8$\times$4) can provide more accurate RAP for each part.
Nevertheless, most backbone network layers are shared between the global features and the local part-based features.
Suppose $\{\mathbf{x}_i\}_{i=1}^7$ are the RAP representations for the 7-parts, the local model is also trained with the TriHard loss defined as below
\begin{equation}
\small
\mathcal{L}_{part} = \mathcal{L}_{TH}( \mathbf{g}\{F_{i=1:7} [ \mathbf{f}_i(\mathbf{x}_i) ]\} ),
\end{equation}
where $\mathbf{f}_i(\cdot)$ is local transformation (i.e., FC-layer) for each part, $F_{i=1:7}[\cdot]$ is a function to aggregate 7-parts information together, $\mathbf{g}\{\cdot\}$ is global transformation (i.e., FC-layer), and $\mathcal{L}_{TH}$ is TriHard loss defined in \autoref{eq:loss_bh}. Note the global transformation $\mathbf{g}\{\cdot\}$ outputs a 107-dimensional feature vector as final representation, similar to the global representation as in previous two baselines.
There are two variants on aggregating local features from 7-parts: \textit{PPbM-a} and \textit{PPbM-b}.
(1) PPbM-a adopts a concatenating function $F_{i=1:7}[\cdot]$ to concatenate features from 7-parts together, as shown in top-right of \autoref{fig:star_model}.
(2) PPbM-b adopts a soft-attention strategy to combine 7-parts together:
\begin{equation}
\small
F_{i=1:7}[\mathbf{y}_i] = \sum\nolimits_{i=1}^7 {\alpha_i \mathbf{y}_i},
\end{equation}
where $\mathbf{y}_i$ is local transformation result for part-$i$,
$\alpha_i$ is soft-attention coefficient obtained similar as Squeeze-excitation networ (SENet) \cite{senet} with shared FC layers as shown in bottom-right of \autoref{fig:star_model}.

Both global representation and pose-part based representation could be trained either with cross-entropy loss or triplet loss.
In our implementation, we defined a combined triplet loss to train the whole network together as
\begin{equation}
\small
    \mathcal{L} = \mathcal{L}_{TH} + \lambda \mathcal{L}_{part},
\end{equation}
where $\lambda$ is a hyper-parameter to control contribution of global and part based representation, with default value $\lambda=1$ .

\section{Experiments}
\subsection{Training Settings}
For each of the three modules (object detection, pose estimation, and tiger re-ID), we choose the most widely used methods either with available open-source code or through re-implementation. Hyper-parameters for these methods are kept as default or recommended except when explicitly noted. We modified the number of training epochs according to dataset size. Specifically, we continued training until the accuracy on the validation set converged. We then re-trained the model on the entire training set, including the validation set, using the same number of epochs.
Since the bounding boxes of tiger are usually horizontal major, which is different from the vertical major case like pedestrian re-ID, we exchange all the corresponding hyper parameters about width and height.
\vspace{-1ex}
\subsection{Benchmark Results}
\subsubsection*{Tiger Detection}

Since the object detector will be deployed on a battery-powered edge camera, we only tested lightweight object detectors to benchmark. In particular, MobileNetv1 \cite{mobilev1} and MobileNetv2 \cite{mobilev2} are widely used image classification models that also serve as backbones for object detector models such as Single Shot Detection (SSD) \cite{ssd}.
This paper benchmarks performance for both SSD-MobileNet-v1 and SSD-MobileNet-v2 models on the detection dataset. We used ImageNet pre-trained backbones for the MobileNet-SSD models. We also benchmarked other efficient object detectors TinyDSOD \cite{tinydsod} and YOLOv3 \cite{yolo3}, in which TinyDSOD is trained from scratch on the training set, while YOLOv3 pre-trains its backbone DarkNet on ImageNet and then strictly followed the official training process.

\begin{figure}[h]
\centering
\small
    \subfigure[Precision-Recall, IOU\textgreater0.5]{
    \includegraphics[width=0.38\linewidth]{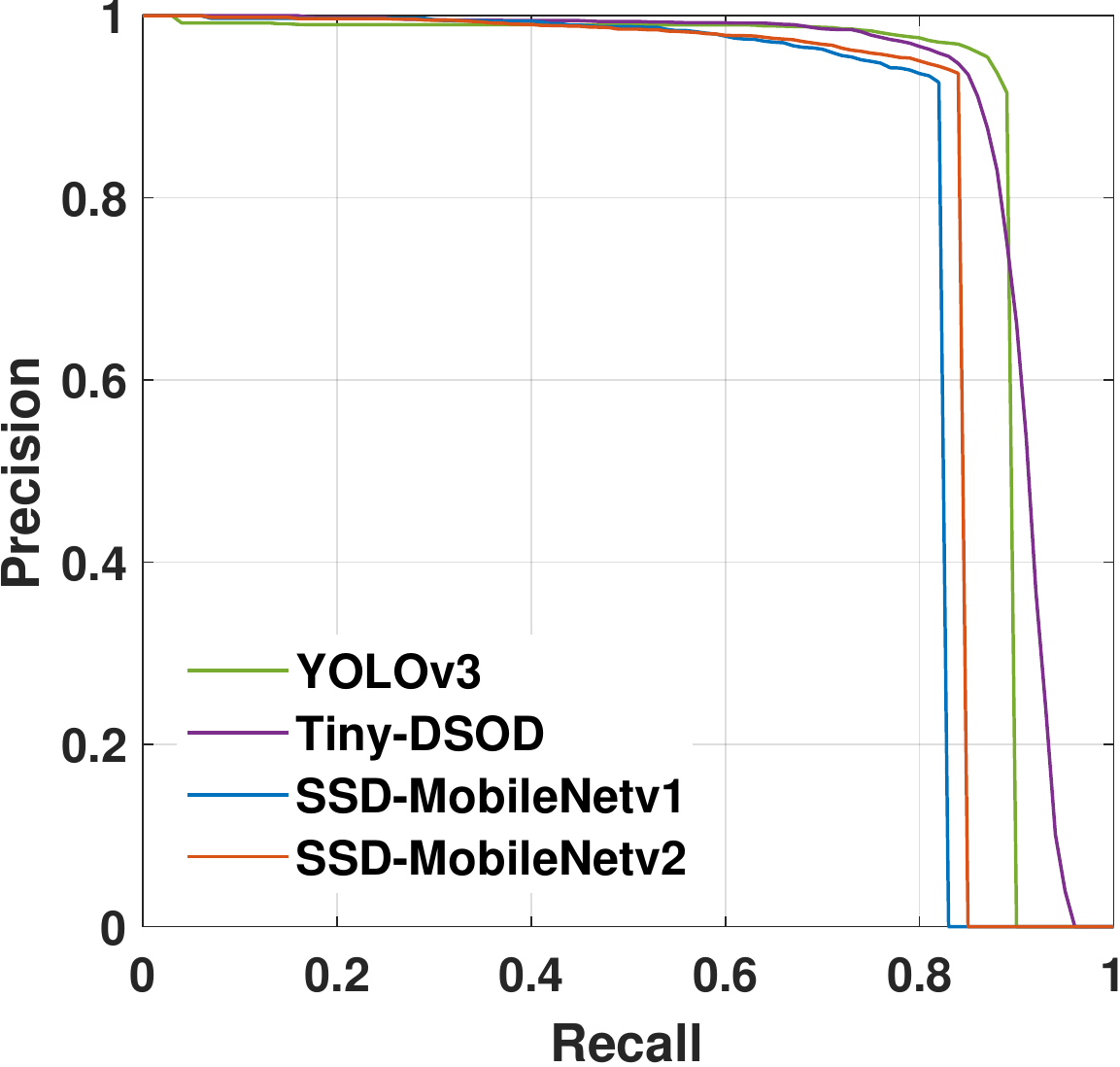}
    }
    \hspace{0.06\linewidth}
    \subfigure[Precision-Recall, IOU\textgreater0.75]{
    \includegraphics[width=0.38\linewidth]{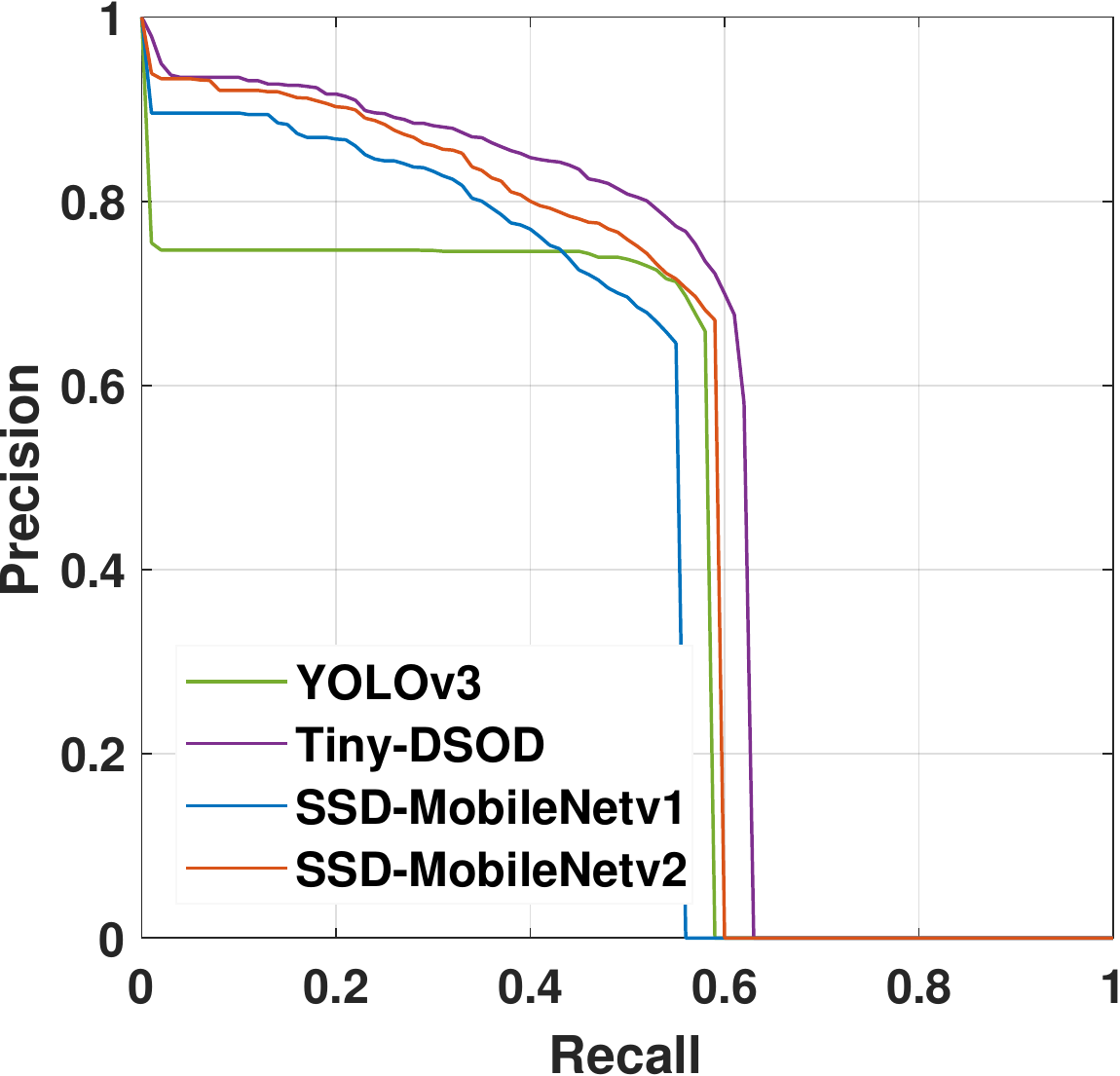}
    }
    \caption{Precision-Recall curve under different IOU thresholds.\label{fig:det_pr}}
    \vspace{-2ex}
\end{figure}

\autoref{fig:det_pr} shows the Precision-Recall curve under different IOU thresholds.
Note $300\times 300$ is the default resolution, and we expect that higher resolutions may provide better results.
{Based on these results, Tiny-DSOD performs best across all metrics.
This may be due to the fact that training from scratch helps the model avoid learning bias or domain difference to better fit our ATRW dataset.}

\begin{table}[h]
    \centering
    \caption{Performance of detectors. mAP is averaged among IOU=0.50:0.05:0.95.}
    \label{tab:det_map}
    \small
    \begin{tabular}{l|c c c c}
    \hline
    Detector          &  mAP  & FLOPs & parameters & {PPF}\\
    \hline
    SSD-MobileNetv1   & {0.446} & 1.2B  & 6.8M & 0.0133\\
    SSD-MobileNetv2   & 0.473 & 1.25B & 14.8M & 0.0344\\
    Tiny-DSOD          & {0.511} & 1.1B  & 0.95M & 0.0736\\
    YOLOv3            & {0.464} & 18.7B & 41.3M & 0.0008\\
    \hline
    \end{tabular}
\end{table}

\subsubsection*{Tiger Pose Estimation}

\begin{figure*}[t]
    \centering
    \subfigure[plain single-cam]{
    \includegraphics[width=0.23\linewidth]{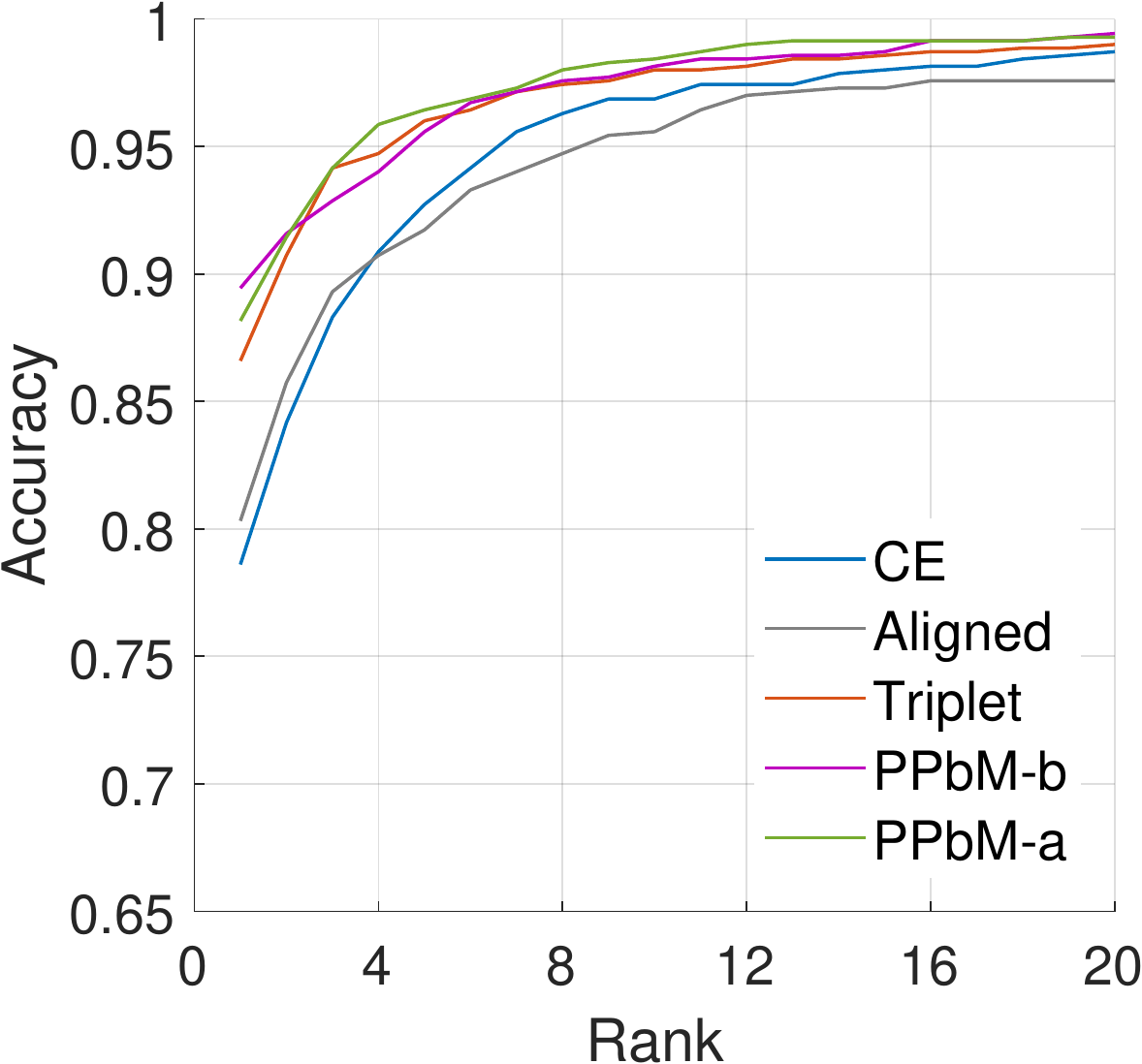}
    }
    \hspace{0ex}
    \subfigure[plain cross-cam]{
    \includegraphics[width=0.23\linewidth]{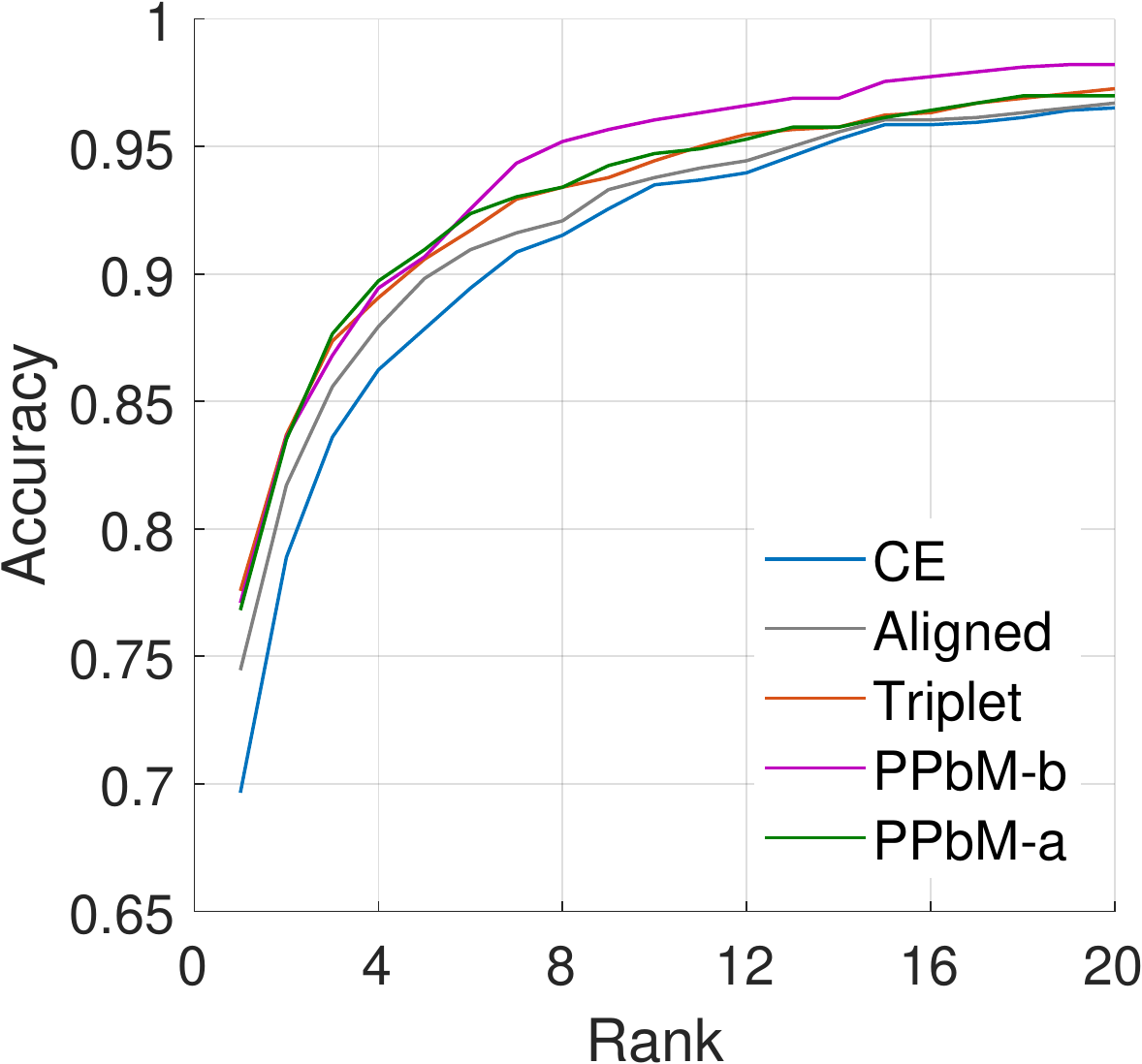}
    }
    \hspace{0ex}
    \subfigure[wild single-cam]{
    \includegraphics[width=0.23\linewidth]{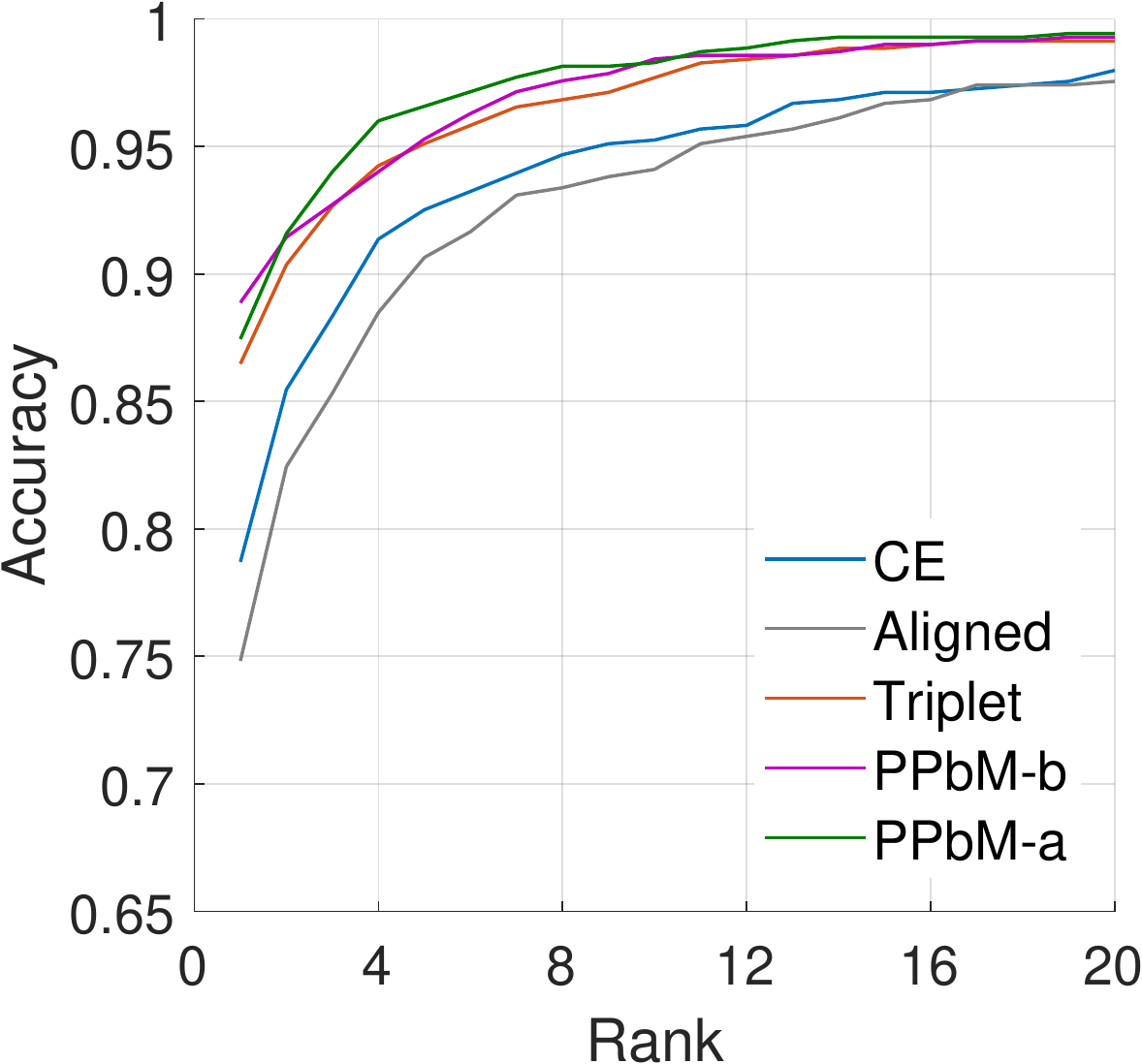}
    }
    \hspace{0ex}
    \subfigure[wild cross-cam]{
    \includegraphics[width=0.23\linewidth]{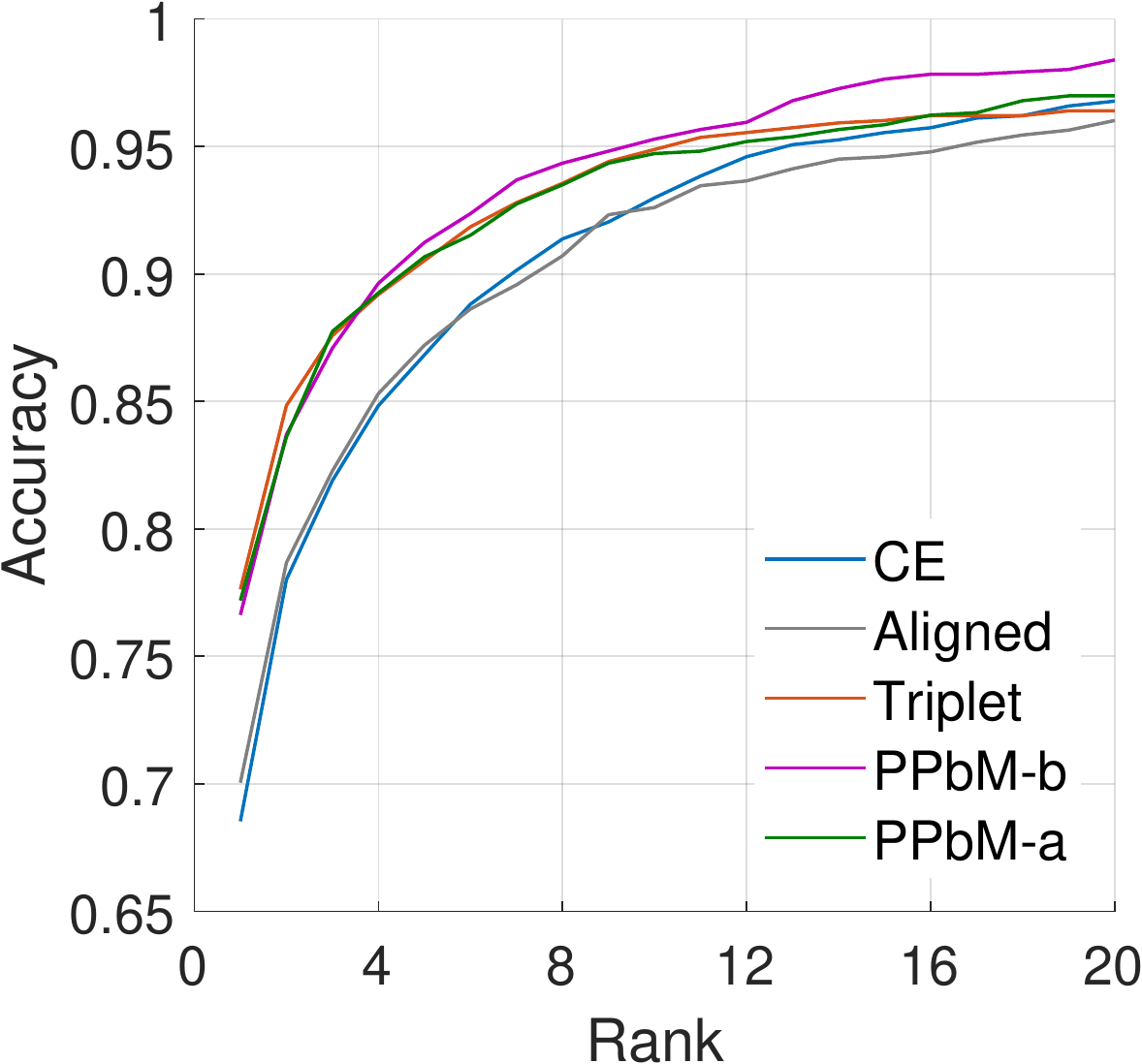}
    }
    \vspace{-2ex}
    \caption{CMC curves for plain and wild tiger re-ID.}\label{fig:reid_metric}
    \vspace{-2ex}
\end{figure*}
\begin{figure}[h]
    \centering
    \subfigure[AP Curve]{
        \includegraphics[width=0.4\linewidth]{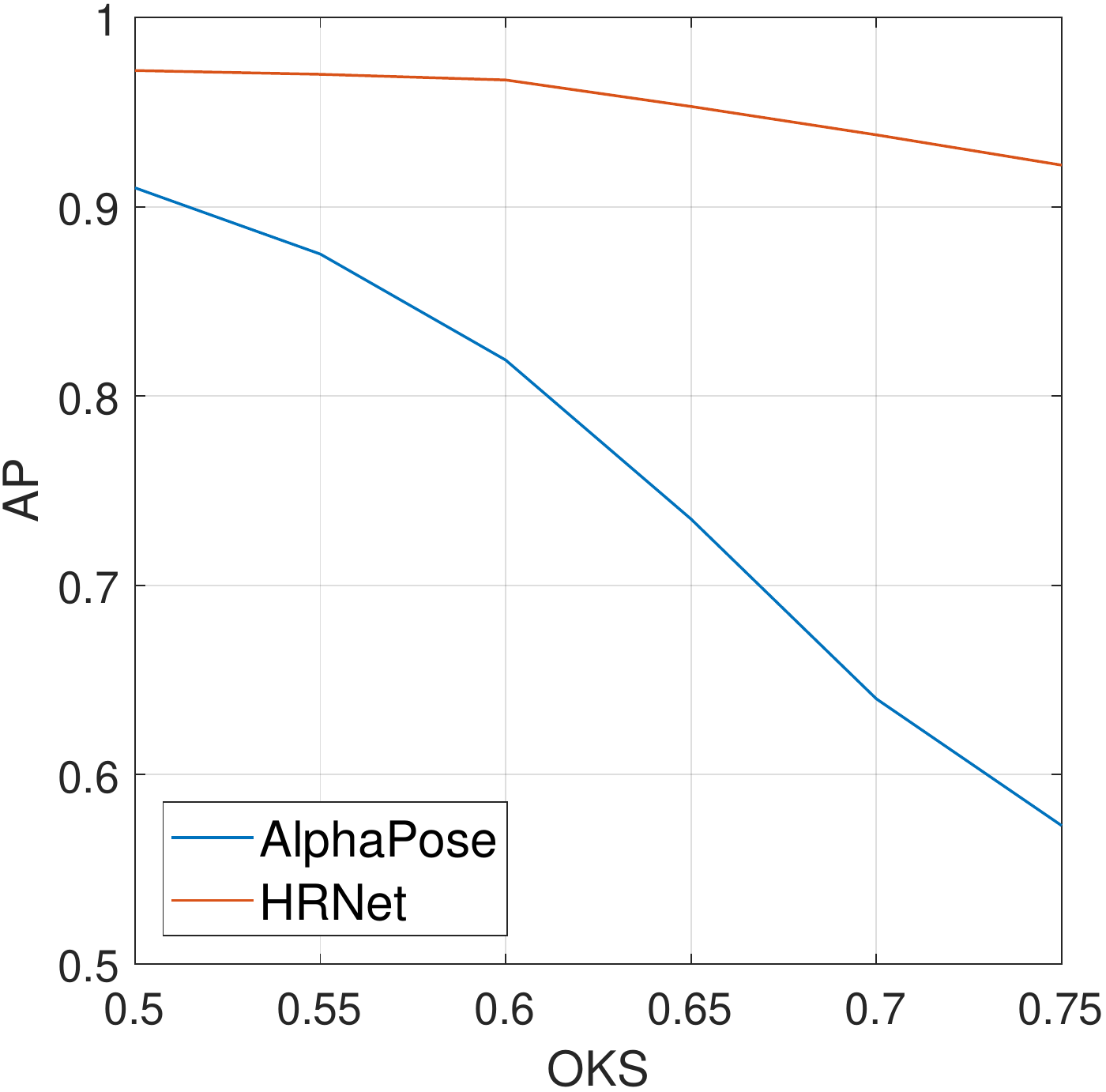}}
    \subfigure[AP Curve]{
        \includegraphics[width=0.4\linewidth]{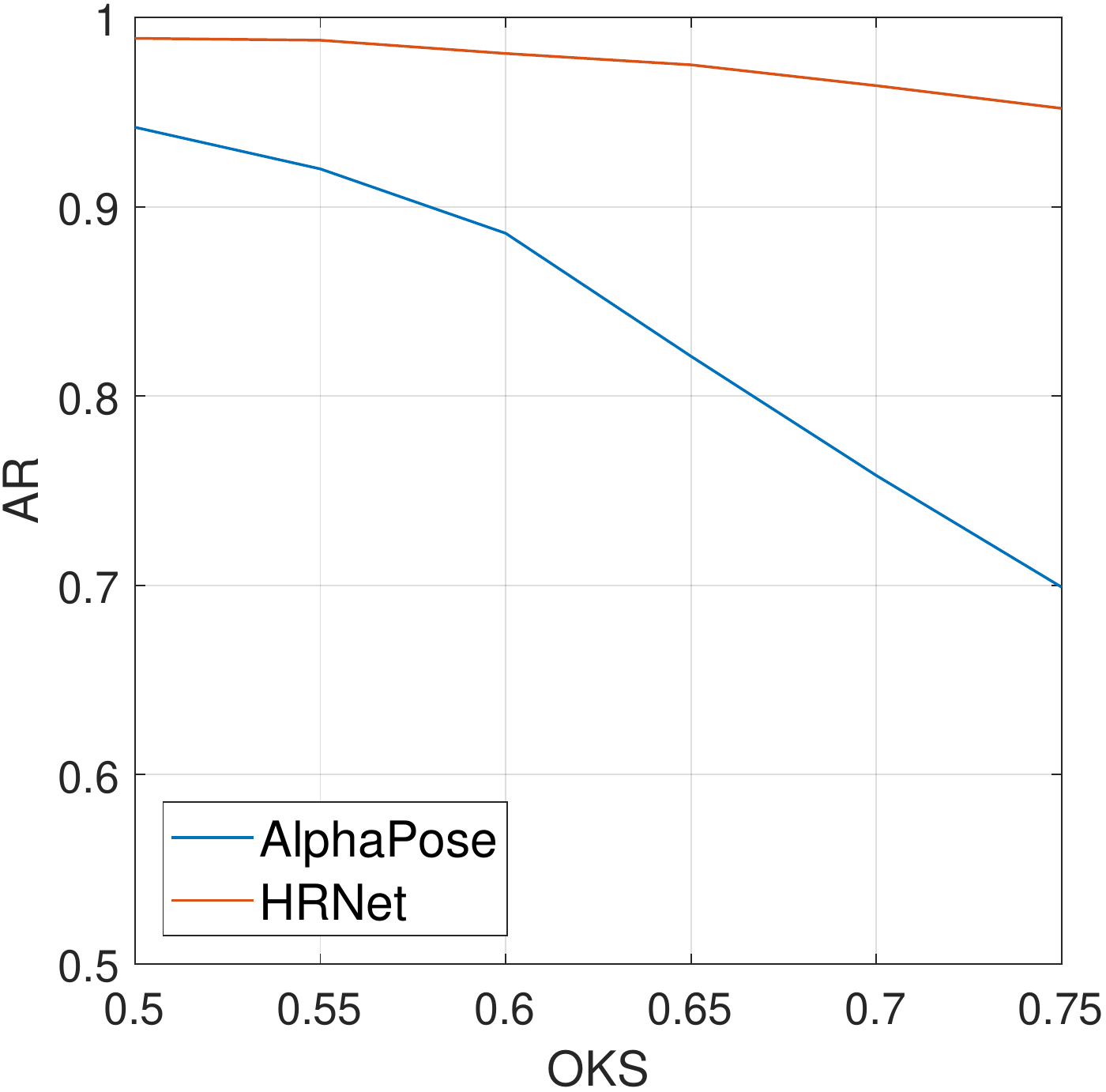}}
    \caption{Curve of AP, AR w.r.t different OKS thresholds.}\label{fig:pose_map}
\end{figure}

\vspace{-1ex}
For the pose estimation, we adopt OpenPose \cite{openpose}, AlphaPose \cite{alphapose} and {HRNet \cite{hrnet}} as they are open-sourced with state of the art results.
Tigers have a very different skeleton definition from humans so we have to modify the original code for human pose accordingly to fit the skeleton definition of tigers. The modification of OpenPose code failed, yielding a non-convergent training procedure.
Fortunately, we successively modified the code of AlphaPose and HRNet, and fine-tuned those pose estimators from provided checkpoint to accommodate the tiger pose configuration.
\autoref{tab:pose_metric} lists the quantitative results by {AlphaPose and HRNet}, and \autoref{fig:pose_map} shows the Average-Precision (AP) and Average-Recall (AR) curve w.r.t different OKS thresholds.
This benchmark reveals that state-of-the-art pose estimator by HRNet can provide fairly accurate (86.9\% for tiger pose versus 77.0\% for human pose \cite{hrnet}) pose results for the wild re-ID purpose. The results can be further improved with more training data.

\begin{table}[h]
    \centering
    \caption{Quantitative result of pose estimators, OKS=0.5.}\label{tab:pose_metric}
    \vspace{-1ex}
    \begin{tabular}{c|c c}
    \hline
    Method     & AP(\%) & AR(\%)\\
    \hline
    OpenPose   & Fail & Fail \\
    AlphaPose  & 0.574 & 0.671 \\
    HRNet      & 0.869 & 0.905 \\
    \hline
    \end{tabular}
\end{table}

\subsubsection*{Plain and Wild Re-ID}

We evaluated both wild case and plain case of tiger re-ID on baseline methods as described in \autoref{sec:baseline}.
In this study, we use ResNet-50~\cite{resnet} pre-trained on ImageNet as the backbone network.
For the plain case, tigers are normalized to $256\times 128$ with manual annotated bounding-box
, while for the wild case, tigers are normalized from automatic bounding-box and pose keypoints. Note that as the final detection result is not high enough, we adopt the trick proposed in \cite{prw}, which utilizes the top-$K$ ($K=10$ at most) scored anchor boxes in the analysis.

\autoref{tab:result} lists the mAP and top-$k$ ($k=1,5$) results for all the compared baseline methods. For the wild case, we evaluate PPbM with bounding boxes provided by SSD-MobileNet-v2, and pose provided by the HRNet.
\autoref{fig:reid_metric} further illustrates the recognition rate vs rank through the Cumulative Match Curve (CMC).
\autoref{fig:reid_metric} further illustrates the recognition rate vs rank through the Cumulative Match Curve (CMC).

\begin{table}[h]
    \centering
    \caption{Benchmark results of baseline re-ID methods on plain and wild re-ID tracks.}\label{tab:result} 
    \footnotesize
    \begin{tabular}{l|l|ccc|ccc}
    \hline
    \multirow{2}{*}{Setting} & \multirow{2}{*}{Method} &\multicolumn{3}{c|}{Single-Cam} &    \multicolumn{3}{c}{Cross-Cam} \\
    & & mAP& top-1 & top-5 & mAP & top-1 & top-5 \\
    \hline
    \multirow{5}{*}{Plain}
    &CE                 & 59.1 & 78.6 & 92.7 & 38.1 & 69.7 & 87.8 \\
    &Triplet loss       & 71.3 & 86.6 & 96.0 & 47.2 & \bf{77.6} & 90.6 \\
    &Aligned-reID       & 64.8 & 81.2 & 92.4 & 44.2 & 73.8 & 90.5 \\
    &PPbM-a (ours)     & \bf{74.1} & 88.2 & \bf{96.4} & \bf{51.7} & 76.8 & \bf{91.0} \\
    &PPbM-b (ours)    & 72.8 & \bf{89.4} & 95.6 & 47.8 & 77.1 & 90.7 \\
    \hline
    \multirow{5}{*}{Wild}
    &CE            & 58.8 & 78.7 & 92.5 & 34.5 & 68.5 & 86.8 \\
    &Triplet loss  & 70.7 & 86.5 & 95.1 & 45.2 & \bf{77.6} & 90.5 \\
    &Aligned-reID  & 58.7 & 74.8 & 90.7 & 41.0 & 70.1 & 87.2 \\
    &PPbM-a (ours) & \bf{71.0} & 87.4 & \bf{96.6} & \bf{50.3} & 77.2 & 90.7 \\
    &PPbM-b (ours) & 69.2 & \bf{88.9} & 95.3 & 46.2 & 76.6 & \bf{91.2} \\
    \hline
    \end{tabular}
\end{table}

We have several observations from the results.
\textit{First}, each method clearly performs much better in the plain case than in the wild case, especially on the cross-camera scenario. This indicates that there is still large improving space for detection and pose estimation modules.
\textit{Second}, PPbM model clearly outperforms other baseline models on both the plain case and the wild case. This indicates that precise pose modeling is important to target with large pose variations like tigers.
\textit{Third}, PPbM-a performs better than PPbM-b in terms of the mAP metric, but performs worse on the top-1 metric mostly. And more interestingly, PPbM-b  outperforms PPbM-a on the CMC curve when $rank > 5$ for the cross-camera case (\autoref{fig:reid_metric}).
We believe that PPbM-b is still powerful and may have potential improving space.
\textit{Forth}, PPbM-a drops 1.4\% in mAP from plain case to wild case, while the Triplet loss baseline drops more than 2.0\%.  This verifies the generalization power for PPbM to some extend.
\textit{Fifth}, there is large improved space for the cross-camera setting, which ensures great value of our ATRW dataset for re-ID research.

\section{Conclusion}

We present a new large-scale wildlife re-ID dataset named ATRW, which contains bounding box, pose keypoint and ID annotations of Amur tigers from multiple wild zoos. Compared to person or vehicle re-ID datasets, wildlife re-ID has a number of novel challenges for re-ID, such as varied pose, lighting, and background environments. In particular, large pose variations due to non-rigid motion require precise target modeling, which are less studied in current re-ID datasets and research. Through systematic benchmarking, we demonstrate that state-of-the-art algorithms are challenged by this dataset, compared to performance on pedestrian or vehicle datasets, and introduce a novel pose part-based model (PPbM) that has significant accuracy gains.
The dataset also expands both the application area and the research challenges for computer vision techniques like re-ID into the important application domain of wildlife conservation.
\begin{acks}
We would like to thank MakerCollider for capturing the raw data and donating it to us for academic usages, and thank WWF Amur tiger and leopard conservation programme team for great support during the whole project. Thanks Paul Lu, Kacie Zhang, Honggang Li, Yunpeng Song, Jiawei Li for numerous helps.
\end{acks}
\newpage
\bibliographystyle{ACM-Reference-Format}
\balance
\bibliography{egbib}

\end{document}